\documentclass{article}

\usepackage{PRIMEarxiv}

\usepackage[utf8]{inputenc} 
\usepackage[T1]{fontenc}    
\usepackage{hyperref}       
\usepackage{url}            
\usepackage{booktabs}       
\usepackage{amsfonts}       
\usepackage{nicefrac}       
\usepackage{microtype}      
\usepackage{lipsum}
\usepackage{fancyhdr}       
\usepackage{graphicx}       
\graphicspath{{media/}}     
\usepackage{multirow}
\usepackage{amsmath}
\usepackage{cleveref}
\usepackage{pifont}
\usepackage{xcolor}
\usepackage{colortbl}

\usepackage{natbib}
\definecolor{mypink}{rgb}{.99,.91,.95}
\definecolor{myblue}{rgb}{0.21,0.49,0.74}

\hypersetup{
hidelinks,
colorlinks=true,
linkcolor=red,
citecolor=myblue
}

\title{From Frames to Sequences: Temporally Consistent Human-Centric Dense Prediction}

\author{
Xingyu\,Miao$^{1}$\;\;\,
Junting\,Dong$^{2,*,\dagger}$\;\;\,
Qin\,Zhao$^{2,3}$\;\;\,
Yuhang\,Yang$^{4}$\;\;\,
Junhao\,Chen$^{5}$\;\;\,
Yang\,Long$^{1,*}$
\\
$^{1}$Durham University \quad $^{2}$Shanghai AI Lab \quad $^{3}$Zhejiang University \quad $^{4}$USTC \quad $^{5}$Tsinghua University\\
Project page: \url{https://xingy038.github.io/F2S/}
}

\begin{document}
\maketitle

\begingroup
\renewcommand{\thefootnote}{\fnsymbol{footnote}}
\footnotetext[2]{Project lead.}
\footnotetext[1]{Corresponding author.}
\renewcommand\thefootnote{}\footnotetext{This work was completed during an internship at Shanghai AI Lab.}
\endgroup

\begin{abstract}
In this work, we focus on the challenge of temporally consistent human-centric dense prediction across video sequences. Existing models achieve strong per-frame accuracy but often flicker under motion, occlusion, and lighting changes, and they rarely have paired human video supervision for multiple dense tasks.  We address this gap with a scalable synthetic data pipeline that generates photorealistic human frames and motion-aligned sequences with pixel-accurate depth, normals, and masks. Unlike prior static data synthetic pipelines, our pipeline provides both frame-level labels for spatial learning and sequence-level supervision for temporal learning. Building on this, we train a unified ViT-based dense predictor that (i) injects an explicit human geometric prior via CSE embeddings and (ii) improves geometry-feature reliability with a lightweight channel reweighting module after feature fusion. Our two-stage training strategy, combining static pretraining with dynamic sequence supervision, enables the model first to acquire robust spatial representations and then refine temporal consistency across motion-aligned sequences. Extensive experiments show that we achieve state-of-the-art performance on THuman2.1 and Hi4D and generalize effectively to in-the-wild videos.
\end{abstract}

\section{Introduction}
In recent years, human-centric vision has advanced in both 2D and 3D applications \citep{xiu2022icon,weng2022humannerf,zhang2023adding,hu2024animate,khirodkar2024sapiens,drobyshev2022megaportraits, zhang2019pose2seg, lin2014microsoft}. Current methods can estimate human pose \citep{cao2017realtime}, and predict dense maps such as depth \citep{khirodkar2024sapiens,saleh2025david} and surface normals \citep{khirodkar2024sapiens,saleh2025david,saito2020pifuhd,xiu2023econ}. Despite recent progress, achieving accurate and temporally consistent predictions in unconstrained videos remains difficult. The main challenges are: (i) the lack of large-scale human-centric video data with paired annotations for dense predictions such as depth, surface normals, and segmentation masks; and (ii) the difficulty for models to simultaneously achieve temporal stability and multi-task learning.

More recently, several methods have shown strong single-image results in estimating depth, surface normals, and segmentation masks from a single image. However, most of these approaches remain optimized for per-frame accuracy and rarely introduce explicit temporal constraints when applied to video. As a result, their predictions often suffer from temporal inconsistency, manifesting as flickering or abrupt discontinuities across frames. For instance, DAViD \citep{saleh2025david} uses post-processing to mitigate flickering, but artifacts persist under fast motion, occlusion, and lighting changes. VDA \citep{chen2025video} achieves temporally consistent depth estimation, due to it is trained on general-purpose datasets, it struggles to reconstruct fine-grained human geometry, including hair strands or clothing wrinkles. Jointly predicting depth and surface normals is also challenging. Although these presentations are geometrically related, their supervision emphasizes different spatial scales, which can destabilize shared representations in multi-task learning. Furthermore, current baseline models are typically trained without human-centric priors, which leads to limited modeling of human structure. The absence of paired human video annotations that simultaneously provide segmentation masks, depth, and surface normals makes it difficult to learn shared features that generalize reliably across tasks.

In this work, we address these issues from both data and modeling perspectives to achieve temporally consistent and multi-task human-centric dense prediction. Instead of relying on a generic model design, we scale up human-centric data and introduce a human-tailored architecture that can effectively exploit this supervision. For data, we propose a human-centric data synthesis pipeline that generates photorealistic images with high-fidelity ground-truth annotations. Beyond static renderings, we incorporate AMASS \citep{AMASS} to produce dynamic sequences with motion-aligned temporal annotations. Each synthesized sample provides static RGB frames with masks, depth, and surface normals, together with dynamic sequences for temporal supervision. For the model, unlike prior works such as Sapiens \citep{khirodkar2024sapiens} and DAViD \citep{saleh2025david}, which primarily scale data on generic architectures, we design a model that explicitly leverages human-centric priors (\textit{i.e.}, CSE \citep{cse}). Our model supports multiple temporally consistent dense prediction tasks, including segmentation, depth, and surface normals, within a single architecture and without task-specific fine-tuning. Trained solely on synthetic data, it achieves state-of-the-art results across benchmarks and generalizes effectively to in-the-wild human images and videos. Our contributions are summarized as follows.
\begin{itemize}
  \item We build a scalable data synthesis pipeline for human-centric frames and videos with pixel-accurate depth, normals, and segmentation. We will release it to support community research on temporal consistency and multi-task learning.
  
  \item We introduce a ViT-based architecture that integrates human geometry priors to jointly predict temporally consistent segmentation, depth, and surface normals. 

  \item To alleviate artifacts arising from feature fusion, we propose an adaptive channel re-weighting module that enhances the reliability of geometry representations.

  \item The method achieves state-of-the-art results on THuman2.1 and Hi4D for both depth and surface normal estimation, and transfers well to in-the-wild videos.
  
\end{itemize}

\section{Related Work}
\subsection{Human vision data}
Recent progress in computer vision largely depends on the availability of high-quality training data \citep{yang2024depth,yang2024depthv2,simeoni2025dinov3,miao2025towards}, and this also applies to human-centric applications \citep{khirodkar2024sapiens}. Tasks such as face detection \citep{viola2004robust}, pose estimation \citep{andriluka20142d}, landmark localization \citep{zhu2012face}, and semantic segmentation \citep{kirillov2023segment} rely on existing annotation tools. In contrast, dense prediction tasks such as depth \citep{wang2025moge} and surface normal \citep{ye2024stablenormal} estimation remain difficult to annotate manually. To address this challenge, several works use multi-view capture \citep{yin2023hi4d,Thuman,martinez2024codec} to reconstruct human meshes. These datasets provide useful supervision, but they show limited subject and scene diversity due to high acquisition costs, and they often lose fine-scale details because they rely on model fitting or photogrammetry. More recently, DAViD \citep{saleh2025david} combines data generation strategies with updated facial models to produce realistic human datasets with precise ground-truth annotations. However, even with large-scale datasets, most of these data are static, and data for dense dynamic prediction is still scarce. Our data synthesis pipeline directly targets this gap and enables high-fidelity synthesis for dynamic scenarios.

\begin{figure*}[t]
    \centering
    \includegraphics[width=0.9725\linewidth]{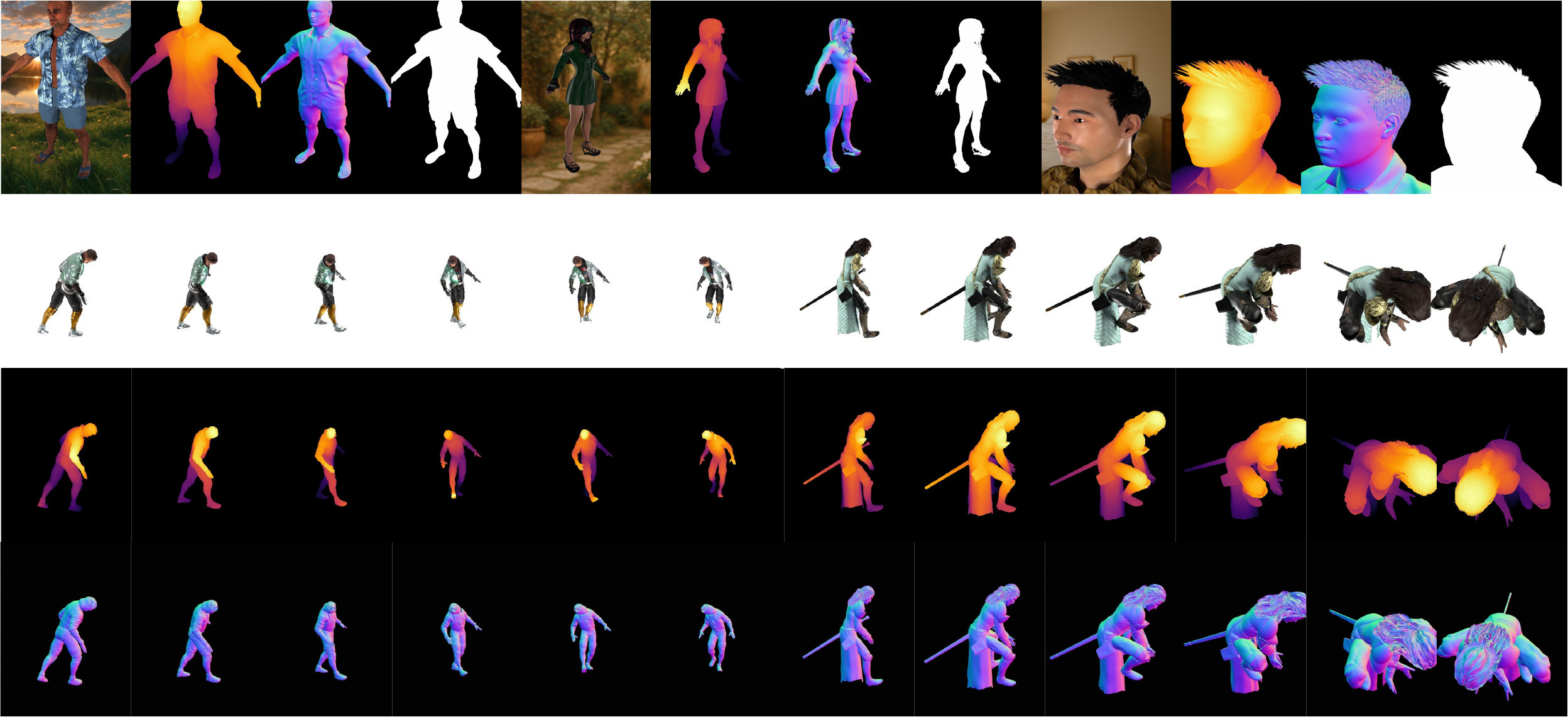}
    \caption{\textbf{Synthesis Data sample.} Ground-truth synthetic annotations of depth, surface normals, and masks.}
    \label{fig:data_sample}
    \vskip -1em
\end{figure*}

\subsection{Human vision Task}
Early research focused primarily on tasks such as human keypoint estimation \citep{chen2018cascaded,fang2017rmpe,huang2017coarse,khirodkar2021multi,newell2016stacked,papandreou2017towards,sun2019deep,xiao2018simple} and body-part segmentation \citep{xia2017joint,xia2016zoom,luo2018macro,gong2018instance,gong2017look,fang2018weakly}. Representative methods such as OpenPose \citep{8765346} tackled multi-person 2D pose estimation. By jointly modeling body, hand, and facial joints, they achieved strong performance in pose and part detection on static images. Recent work has expanded to broader dense prediction tasks beyond keypoints and segmentation, including depth estimation \citep{bhat2023zoedepth,yin2023metric3d,jafarian2021learning,birkl2023midas} and surface normal prediction \citep{eigen2015predicting,ladicky2014discriminatively,saito2020pifuhd,xiu2023econ}. For example, Sapiens \citep{khirodkar2024sapiens} leverages large-scale in-the-wild human images for pre-training and fine-tuning on 2D pose estimation, part segmentation, depth, and normal prediction, showing strong generalization to natural scenes. DAViD \citep{saleh2025david} further achieves competitive results by fine-tuning DINOv2 \citep{oquab2023dinov2} on synthetic data. Despite their broad task coverage, these methods remain limited in stability when applied to dynamic video scenes. In this work, we go beyond static image training by introducing supervisory signals from video sequences. This improves stability under motion, occlusion, and illumination variations, leading to more robust and generalizable predictions in natural scene videos.

\subsection{Dense Prediction Architectures}
Dense prediction has transitioned from CNN encoder–decoder baselines \citep{ronneberger2015u,chen2018encoder} with skip connections to transformer backbones trained with strong pretraining and scalable supervision. DPT \citep{ranftl2021vision} shows that a ViT encoder \citep{dosovitskiy2020image} with a lightweight convolutional decoder yields fine-grained and globally consistent outputs for depth and segmentation, and it generalizes well across datasets. Large-scale self-supervised pretraining further improves transfer, and features from DINOv2 \citep{oquab2023dinov2} are widely used as a shared backbone for dense tasks without heavy task-specific heads. The Depth Anything family, especially Depth Anything V2 \citep{yang2024depthv2}, scales supervision using a stronger synthetic teacher and large pseudo-labeled collections of real images. The models span tens of millions to over one billion parameters and achieve improved accuracy and speed. Marigold \citep{ke2023repurposing} adapts a pretrained latent diffusion model to monocular depth with lightweight fine-tuning on synthetic data and reports strong cross-dataset results. For human-centric dense estimation, Sapiens \citep{khirodkar2024sapiens} uses a ViT backbone with lightweight task heads. In contrast, DAViD \citep{saleh2025david} employs a dual-branch design with a ViT encoder branch and a shallow fully convolutional branch, and the features are fused in a DPT-style decoder before lightweight heads. We propose a model that injects explicit human priors into the backbone to encode body topology and part correspondence, which improves human-centric dense prediction.

\section{Methodology}
\label{sec:method}
\subsection{Human-centric Synthetic Data Pipeline}
Our data synthesis pipeline consists of two stages: composition and rendering.

\paragraph{Composition stage.} We leverage some character generation software (i.e., DAZ 3D\footnote{\url{https://www.daz3d.com/}}, MakeHuman\footnote{\url{http://www.makehumancommunity.org/}}, Character Creator \footnote{\url{https://www.reallusion.com/character-creator/}}) to compose clothed human models. Assets are divided into four categories—body, top, bottom, and shoes—so that they can be sampled independently. We randomize body shape and pair tops, bottoms, and shoes to generate diverse outfits. This independent sampling strategy increases the coverage of outfit combinations without requiring manual curation. For asset textures, we apply three categories of augmentations to the diffuse maps. The first introduces appearance variations through hue adjustment, per-channel intensity scaling, and low-magnitude noise. The second generates uniform solid-color textures to diversify simple surface representations. The third replaces textures using external resources, including the Describable Textures Dataset \citep{cimpoi14describing}, the ALOT dataset \citep{burghouts2009material}, and an internal texture collection. For these replacements, we apply preprocessing operations such as resizing, tiling, mirrored tiling, and HSV-based recoloring to accommodate both colored and grayscale inputs. In total, we compose about 200K unique identities for rendering.

\paragraph{Rendering stage.} We import the composed models into Blender \footnote{\url{https://www.blender.org/}}
to render RGB images, depth maps, surface normal maps, and segmentation masks, from which we generate both static and dynamic data (In \Cref{fig:data_sample}).  For image data, we follow the Sapiens protocol \citep{khirodkar2024sapiens} by randomly sampling camera viewpoints to render three perspectives: face, upper body, and full body. For video data, we animate the models with motion capture sequences from the AMASS dataset \citep{AMASS}, which provides skeletal trajectories. We exclude sequences with poses such as lying and uniformly sample up to 500 frames per sequence. Each model is paired with a randomly selected trajectory. To further increase diversity, we randomize the camera focal length, enable subject-tracking, and apply camera rotations around the animated model during rendering.

\begin{figure*}[t]
    \centering
    \includegraphics[width=0.9725\linewidth]{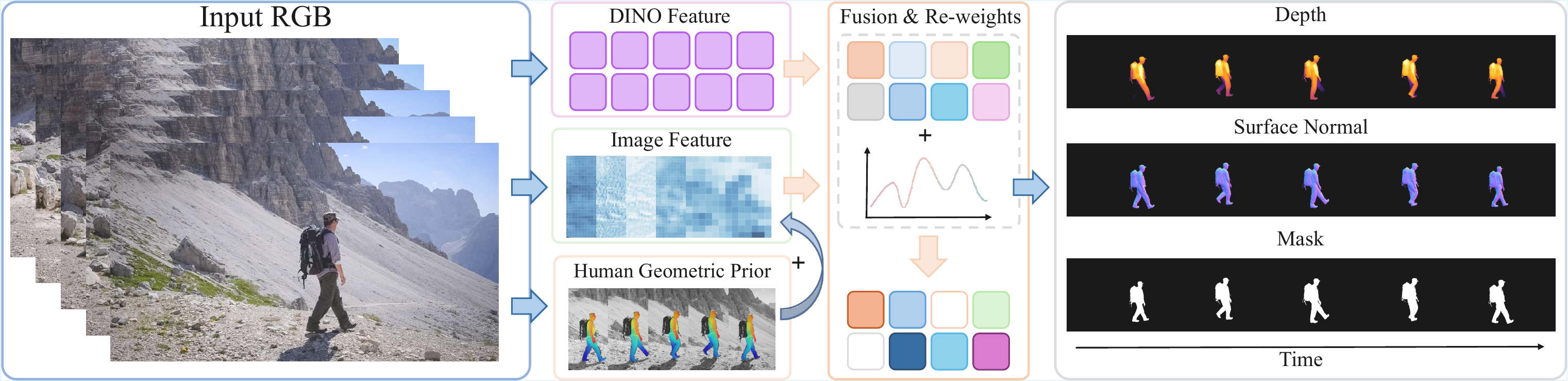}
    \caption{\textbf{Pipeline overview.} Given a sequence of RGB frames, our model extracts DINO features, global image features, and human geometric priors. These features are fused and re-weighted to generate enhanced representations for predicting temporally consistent depth, surface normals, and segmentation masks.}
    \label{fig:pipeline}
    \vskip -1em
\end{figure*}
\subsection{Model Architecture}
We build upon the recent paradigm of ViT-based dense prediction and adapt it to human-centric tasks. While approaches such as Sapiens \citep{khirodkar2024sapiens} and DAViD \citep{saleh2025david} achieve strong performance, they remain largely task-agnostic and do not explicitly incorporate human geometry. Our objective is to introduce human geometric priors into the representation learning process, enabling a framework for human-centric dense prediction tasks (\Cref{fig:pipeline}).

\paragraph{Encoder and Decoder.} We adopt the DINO series \citep{oquab2023dinov2} as the encoder $\mathcal{E}_{\mathrm{DINO}}$, which extracts global representations from the input image $\mathbf{x} \in \mathbb{R}^{H \times W \times 3}$ as $\mathbf{F}_{\mathrm{enc}}=\mathcal{E}_{\mathrm{DINO}}(\mathbf{x}).$ The DPT \citep{ranftl2021vision} decoder $\mathcal{D}_{\mathrm{DPT}}$ transforms this representation into multi-scale features $\mathbf{F}_{\mathrm{DPT}}=\mathcal{D}_{\mathrm{DPT}}(\mathbf{F}_{\mathrm{enc}}).$ On top of these features, we leverage three lightweight task heads $\mathcal{H}$ that produce predictions for depth, surface normals, and foreground/background segmentation $\{\hat{\mathbf{D}},\hat{\mathbf{N}},\hat{\mathbf{S}}\}=\mathcal{H}(\mathbf{F}_{\mathrm{DPT}}).$ To capture the temporal relationship between frames, we inject four temporal blocks $\mathcal{T}$ into the decoder as bridges connecting different frames. The structure of the temporal blocks in the model is similar to that in AnimateDiff \citep{guo2023animatediff} and VDA \citep{chen2025video}, consisting of several temporal attention blocks. 

\paragraph{Local Geometry Enhancement.} While DINO tokens effectively encode semantic information and capture long-range dependencies, they generally lack fine details such as edges and textures. Inspired by the Resizer module in DAViD \citep{saleh2025david}, we introduce a lightweight CNN branch $\mathcal{E}_{\mathrm{CNN}}.$ This branch directly extracts edges and textures from the input image as $\mathbf{F}_{\mathrm{CNN}}=\mathcal{E}_{\mathrm{CNN}}(\mathbf{x}).$ The final fused representation is then obtained by concatenating the decoder features with the CNN features, followed by a nonlinear mapping: $\mathbf{F}_{\mathrm{fusion}}=\phi_{\mathrm{fusion}}([\mathbf{F}_{\mathrm{DPT}},\mathbf{F}_{\mathrm{CNN}}]),$ where $[\,\cdot\,,\cdot\,]$ denotes channel concatenation and $\phi$ denotes a nonlinear mapping. 

\paragraph{Channel Weight Adaptation (CWA).} While the fusion design preserves global semantics and strengthens texture cues, the lightweight CNN branch can introduce redundant appearance signals. DAViD observed similar issues, with appearance details such as tattoos and lighting patterns sometimes being mistaken for geometric shapes. To alleviate this, we introduce a channel weight adaptation module to reweight the channel weights of the fused features. 
Specifically, given the fused feature map $\mathbf{F}_{\text{fusion}} \in \mathbb{R}^{C \times H \times W}$, we introduce a light-weight channel-wise reweighting block to adjust the contribution of each channel.
We first apply global average pooling over the spatial dimensions to obtain a channel descriptor
\begin{equation}
q_c = \frac{1}{HW} \sum_{h=1}^{H} \sum_{w=1}^{W} \mathbf{F}_{\text{fusion}}(c,h,w), \quad c = 1,\dots,C,
\end{equation}
which forms a vector $q \in \mathbb{R}^{C}$.
This vector is then passed through a small two-layer MLP with a non-linear activation and a sigmoid function $\sigma(\cdot)$ to produce per-channel weights
\begin{equation}
a = \sigma(\operatorname{MLP}(q)) \in (0,1)^C.
\end{equation}
Finally, the fused features are rescaled channel-wise as
\begin{equation}
\mathbf{F}'_{\text{fusion}}(c,h,w) = a_c \, \mathbf{F}_{\text{fusion}}(c,h,w),
\end{equation}
where $a_c$ denotes the weight of channel $c$.

The CWA is trained jointly with depth and normal objectives, guiding the network to assign larger weights to channels. In this way, it reduces the weights of texture- and lighting-dominated channels while increasing the weights of geometry-related channels, thereby weakening the influence of appearance information on geometry prediction and maintaining the consistency of global representation.

\paragraph{Human Geometric Prior.} Previous approaches mainly rely on general designs and data-centric scaling (larger and cleaner datasets), which raises the capacity from the data side but leaves model-side priors underused. We therefore inject a human-specific prior to strengthen the representation of the human body structure. A straightforward option is to use DensePose-like UV maps \citep{guler2018densepose} so that the network predicts geometry for different body parts. However, due to the lack of such supervised data and in the multi-task setting, this option usually fails to achieve stable convergence. Instead, we adopt CSE \citep{cse} as a stable geometric prior. Given a human image, the CSE encoder $\mathcal{E}_{\mathrm{CSE}}$ produces continuous geometric embeddings $\mathbf{z}=\mathcal{E}_{\mathrm{CSE}}(\mathbf{x})$, which we fuse with decoder features to impose shape-aware constraints on the predictions. Let $F_{\text{DPT}}$ denote the decoder features. To inject the human geometric prior into the representation, we project $\mathbf{z}$ to the same channel dimension and spatial resolution as $F_{\text{DPT}}$ using a $1 \times 1$ convolution followed by bilinear upsampling, and then fuse it with the decoder features by element-wise addition:
\begin{equation}
\tilde{\mathbf{z}} = \psi(\mathbf{z}) \in \mathbb{R}^{C \times H \times W}.
\end{equation}
The prior is then fused with the decoder features by element-wise addition:
\begin{equation}
\mathbf{F}'_{\text{DPT}} = \mathbf{F}_{\text{DPT}} + \tilde{\mathbf{z}}.
\end{equation}

\subsection{Training Pipeline}
\label{sec:3.3}
To achieve multi-task human-centered temporal consistency, we adopt a two-stage training strategy. In stage 1, the model is pretrained on synthetic image data to learn spatially consistent fundamental representations. In stage 2, we inject the temporal module and continue training on synthetic video data with flow-guided stabilization term to capture temporal information and maintain consistency.

\subsubsection{Stage 1: Static Image Model Training}

\paragraph{Monocular Depth Estimation.}
For depth estimation, given a depth map $\mathbf{d}^*$, we normalize it to the range $[0,1]$ by $\mathbf{d}=\frac{\mathbf{d}^*-\min(\mathbf{d}^*)}{\max(\mathbf{d}^*)-\min(\mathbf{d}^*)}$. Let $\hat{\mathbf{D}}$ be the predicted relative depth. We follow previous work \citep{birkl2023midas} to estimate per-image scale and shift $(s,t)$. The depth loss is:
\begin{equation}
\mathcal{L}_{\mathrm{depth}}
=\lVert s\,\hat{\mathbf{D}}+t-\mathbf{d}\big\rVert_{2}
+\omega_{\mathrm{grad}}\ \mathcal{L}_{\mathrm{grad}}(s\hat{\mathbf{D}}+t,\ \mathbf{d}),
\end{equation}
where $\mathcal{L}_{\mathrm{grad}}$ is the gradient term \citep{hu2019revisiting} to encourage sharp boundaries and local continuity.

\paragraph{Surface Normal Estimation.}
The normal head outputs 3-channels $(x,y,z)$.
Let $\mathbf{N}$ be the ground-truth normal and $\hat{\mathbf{N}}$ the prediction. The base loss combines a $L_1$ term with a cosine term:
\begin{equation}
\mathcal{L}_{\mathrm{base}}
=\lVert \mathbf{N}-\hat{\mathbf{N}}\rVert_{1}
+ \big(1-\mathbf{N}\cdot\hat{\mathbf{N}}\big).
\end{equation}

We observe that when depth and normal heads are trained jointly, the predicted normals often lose fine texture details. This occurs because depth supervision relies on global geometric consistency and largely ignores high-frequency signals that are uninformative for relative depth. Since depth typically converges faster and more stably than normal estimation, it tends to dominate the shared representation during training. Consequently, the learned features emphasize smooth, low-frequency structures while suppressing texture cues, leading to over-smoothed surface normals, particularly on clothing, accessories, and hair. To mitigate this effect and enhance spatial coherence, we introduce an edge-aware gradient loss and a multi-scale Laplacian loss. Let $\nabla$ denote the Sobel operator and $\Delta$ the discrete Laplacian. Define an edge weight using the magnitude of ground-truth normal gradients $w_{\mathrm{edge}}
=1+\eta\ \frac{\lVert\nabla \mathbf{N}\rVert-\min \lVert\nabla \mathbf{N}\rVert}
{\max \lVert\nabla \mathbf{N}\rVert-\min \lVert\nabla \mathbf{N}\rVert}$. The regularizers are:
\begin{equation}
\mathcal{L}_{\mathrm{grad}}^{n}
= w_{\mathrm{edge}}\ 
\big\lVert\nabla (\hat{\mathbf{N}})-\nabla (\mathbf{N})(x)\big\rVert_{1}\quad\mathcal{L}_{\mathrm{lap}}
= w_{\mathrm{edge}}\ 
\big\lVert\Delta (\hat{\mathbf{N}})-\Delta (\mathbf{N})\big\rVert_{1}.
\end{equation}
\begin{figure}
    \centering
    \includegraphics[width=\linewidth]{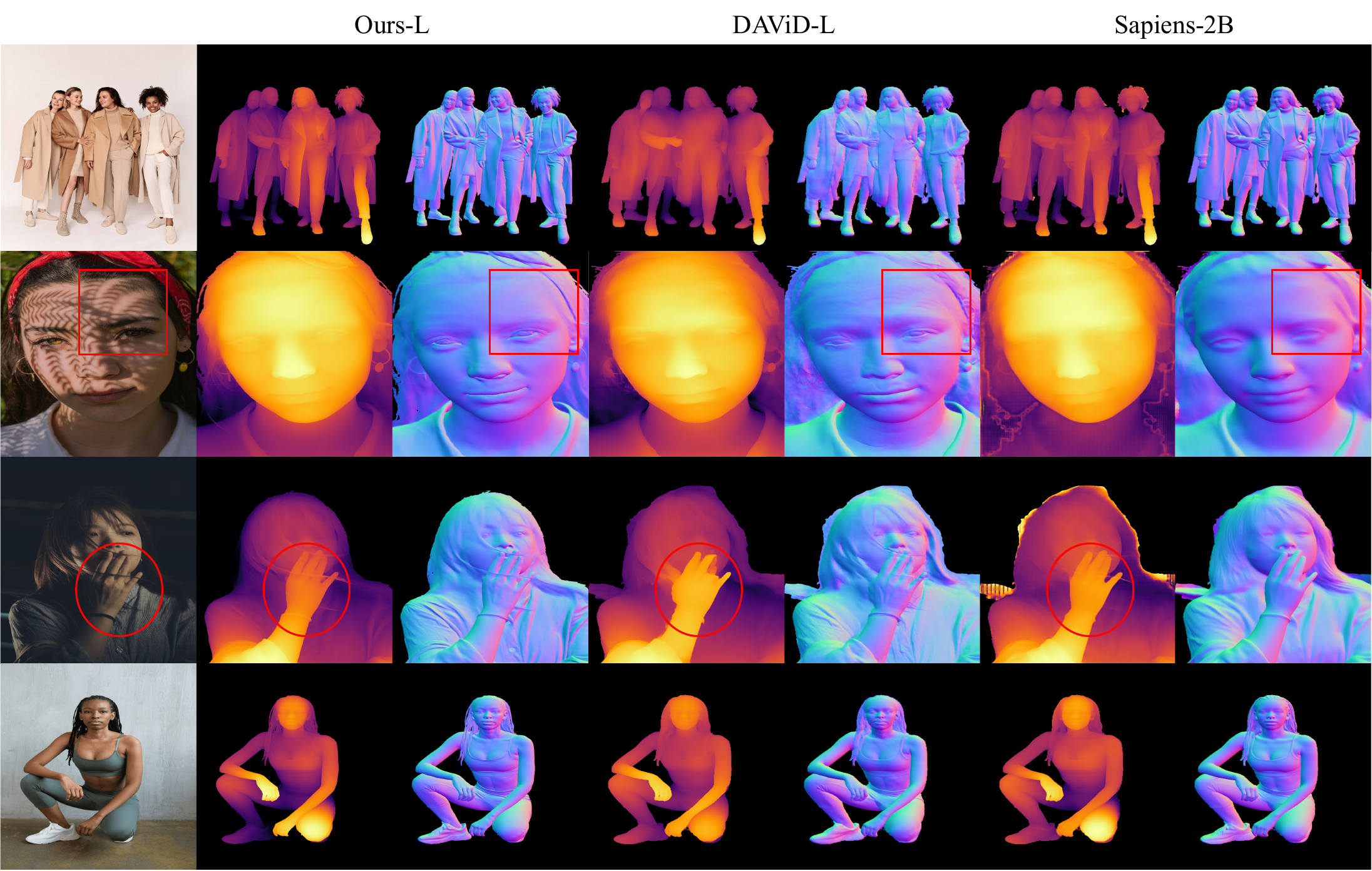}
    \caption{\textbf{Qualitative comparison} on challenging images in the wild.}
    \label{fig:static}
\end{figure}

The surface normal loss is:
\begin{equation}
\mathcal{L}_{\mathrm{normal}}
=\mathcal{L}_{\mathrm{base}}+\alpha\ \mathcal{L}_{\mathrm{grad}}^{n}
+\beta \mathcal{L}_{\mathrm{lap}},
\end{equation}
where $\alpha$ and $\beta$ are regularizers weights.

\paragraph{Foreground Segmentation.}
To provide human-centric foreground guidance for geometry-related tasks, we introduce a lightweight segmentation head that predicts a soft mask $\hat{\mathbf{S}}$ over the human region. Designed as an auxiliary branch that shares the same backbone as the depth and normal heads, this head supplies soft human masks that guide the depth and normal predictors to focus on the foreground and obtain cleaner supervision near human boundaries. The segmentation head predicts a soft mask $\hat{\mathbf{S}}.$ We use binary cross-entropy to supervise them:
\begin{equation}
\mathcal{L}_{\mathrm{seg}}=\mathcal{L}_{\mathrm{BCE}}(\hat{\mathbf{S}},\mathbf{S}),
\end{equation}
and use $\mathbf{S}$ as the mask for depth and surface normal supervision. Finally, the Stage-1 objective is:
\begin{equation}
\mathcal{L}_{\mathrm{stage1}}
=\lambda_{d}\mathcal{L}_{\mathrm{depth}}
+\lambda_{n}\mathcal{L}_{\mathrm{normal}}
+\lambda_{s}\mathcal{L}_{\mathrm{seg}}.
\end{equation}

\subsubsection{Stage 2: Dynamic Video Model Training}
To address frame-to-frame instability in dense video prediction, existing methods can be broadly divided into two categories. The first category is the TGM loss proposed in VDA \citep{chen2025video}, which enforces temporal consistency by constraining the depth gradient between adjacent frames. The second category is flow-based temporal consistency. Since our work involves not only depth but also normal estimation, with supervision primarily focused on human foreground regions, TGM is not directly suitable. It is restricted to depth prediction and tends to weaken supervision on fast-moving or occluded foreground regions. In contrast, flow-based methods explicitly establish correspondences across frames, which enables stable supervision in the foreground and naturally extends to enforcing directional consistency for surface normals.

\begin{table}
\renewcommand{\arraystretch}{1.3}
\setlength{\tabcolsep}{3pt}
\caption{\textbf{Quantitative comparison} for depth estimation on THuman2.1 and Hi4D dataset. Note that the parameter size of Sapiens-0.3B is equivalent to that of large models of ViT-based methods.}
\label{tab:static_depth}
\centering
\begin{tabular}{lcccccccc}
\toprule
\multirow{2}{*}{Methods} 
& \multicolumn{2}{c}{TH2.1-Face} 
& \multicolumn{2}{c}{TH2.1-UpperBody} 
& \multicolumn{2}{c}{TH2.1-FullBody} 
& \multicolumn{2}{c}{Hi4D} 
\\ 
\cmidrule(lr){2-3} \cmidrule(lr){4-5}\cmidrule(lr){6-7}\cmidrule(lr){8-9}
& RMSE $\downarrow$ & AbsRel $\downarrow$ & RMSE $\downarrow$ & AbsRel $\downarrow$ & RMSE $\downarrow$ & AbsRel $\downarrow$ & RMSE $\downarrow$  & AbsRel $\downarrow$
\\ 

\cmidrule(r){1-1} \cmidrule(lr){2-7} \cmidrule(l){8-9}

DA-B & 0.0267& 0.0157& 0.0324& 0.0175& 0.0366& 0.0176& 0.0954& \textbf{0.0251}\\
DA2-B & 0.0328 & 0.0204 & 0.0423 & 0.0241 & 0.0404 & 0.0209 & \underline{0.0930} & \underline{0.0262} \\
MoGe2-B & 0.0274 & 0.0165 & 0.0326 & 0.0179 & 0.0451 & 0.0208 & 0.1104 & 0.0281 \\
DAViD-B & \underline{0.0254} & \underline{0.0147} & \underline{0.0262} & \underline{0.0143} & \underline{0.0304} & \underline{0.0148} & 0.0947& 0.0266\\
\cellcolor{mypink}Ours-B & \cellcolor{mypink}\textbf{0.0193} & \cellcolor{mypink}\textbf{0.0112}& \cellcolor{mypink}\textbf{0.0228} & \cellcolor{mypink}\textbf{0.0126} & \cellcolor{mypink}\textbf{0.0293} & \cellcolor{mypink}\textbf{0.0146} & \cellcolor{mypink}\textbf{0.0928} & \cellcolor{mypink}0.0277\\

\cmidrule(r){1-1} \cmidrule(lr){2-7} \cmidrule(l){8-9}

DA-L & 0.0236 & 0.0138 & 0.0297 & 0.0162 & 0.0323 & 0.0160 & 0.0845 & 0.0228\\
DA2-L & 0.0303 & 0.0187 & 0.0381 & 0.0216 & 0.0379 & 0.0197 & \underline{0.0844} & 0.0239\\
MoGe-L & 0.0222 & 0.0132 & 0.0276 & 0.0145 & 0.0361 & 0.0159 & 0.0915 & \underline{0.0216}\\
MoGe2-L & 0.0231 & 0.0136 & 0.0294 & 0.0154 & 0.0349 & 0.0149 & 0.0892 & \textbf{0.0208}\\
DAViD-L & 0.0256 & 0.0149 & 0.0262 & 0.0144 & 0.0293 & 0.0142 & 0.0889 & 0.0244\\
Sapiens-0.3B & \underline{0.0150} & \underline{0.0089} & \underline{0.0184} & \underline{0.0105} & \underline{0.0239} & \underline{0.0117} & 0.1349 & 0.0412\\
\cellcolor{mypink}Ours-L & \cellcolor{mypink}\textbf{0.0147} & \cellcolor{mypink}\textbf{0.0086} & \cellcolor{mypink}\textbf{0.0174} & \cellcolor{mypink}\textbf{0.0098} & \cellcolor{mypink}\textbf{0.0218} & \cellcolor{mypink}\textbf{0.0110} & \cellcolor{mypink}\textbf{0.0700} & \cellcolor{mypink}\textbf{0.0208}\\ 

\cmidrule(r){1-1} \cmidrule(lr){2-7} \cmidrule(l){8-9}

\color[HTML]{C0C0C0}Sapiens-0.6B & \color[HTML]{C0C0C0}0.0152 & \color[HTML]{C0C0C0}0.0087 & \color[HTML]{C0C0C0}0.0183 & \color[HTML]{C0C0C0}0.0104 & \color[HTML]{C0C0C0}0.0236 & \color[HTML]{C0C0C0}0.0119 & \color[HTML]{C0C0C0}0.1317 & \color[HTML]{C0C0C0}0.0407\\
\color[HTML]{C0C0C0}Sapiens-1B & \color[HTML]{C0C0C0}0.0119 & \color[HTML]{C0C0C0}0.0067 & \color[HTML]{C0C0C0}0.0145 & \color[HTML]{C0C0C0}0.0080 & \color[HTML]{C0C0C0}0.0179 & \color[HTML]{C0C0C0}0.0087 & \color[HTML]{C0C0C0}0.1151 & \color[HTML]{C0C0C0}0.0356\\
\color[HTML]{C0C0C0}Sapiens-2B & \color[HTML]{C0C0C0}0.0112 & \color[HTML]{C0C0C0}0.0061 & \color[HTML]{C0C0C0}0.0156 & \color[HTML]{C0C0C0}0.0086 & \color[HTML]{C0C0C0}0.0172 & \color[HTML]{C0C0C0}0.0082 & \color[HTML]{C0C0C0}0.1060 & \color[HTML]{C0C0C0}0.0327\\ \bottomrule
\end{tabular}
\end{table}

Based on this, we keep all spatial losses and introduce optical-flow-based stabilization. For adjacent frames $k$ and $k{+}1$, we denote the forward and backward flows as $\mathcal{O}_{k\to k+1}$ and $\mathcal{O}_{k+1\to k}$. Warping with flow $\mathcal{O}$ is denoted as $\mathcal{W}(\cdot,\mathcal{O})$. To ensure reliable correspondences, we further apply a cycle-consistency mask $\mathcal{M}_{\mathrm{cyc}}
=\mathbf{1}\Big(\lVert \mathcal{O}_{k\to k+1}(\mathcal{O}_{k+1\to k})-\mathbf{x}\rVert_2 \le \tau_c\Big)$.
We also suppress unstable boundary pixels using a non-edge mask from predicted depth edges.
Let $\mathbf{E}_k$ be the edge map extracted from the current predicted depth, and let its dilated form be used to compute the edge mask $\mathcal{M}_{\mathrm{edge}}=1-\mathrm{Dilate}(\mathbf{E}_k)$. The valid set is $\mathcal{M}=\mathcal{M}_{\mathrm{cyc}} \cap \mathcal{M}_{\mathrm{edge}}$. Depth stabilization uses a bidirectional, flow-aligned $L_1$ loss: 
\begin{equation}
\mathcal{L}_{\mathrm{temp}}^{d}
=\frac{1}{|\mathcal{M}|}\,\big\lVert 
\mathcal{M}\odot\big(\hat{\mathbf{D}}_{k}-\mathcal{W}(\hat{\mathbf{D}}_{k+1},\mathcal{O}_{k \to k+1})\big)
\big\rVert_{1}+\frac{1}{|\mathcal{M}|}\,\big\lVert 
\mathcal{M}\odot\big(\hat{\mathbf{D}}_{k+1}-\mathcal{W}(\hat{\mathbf{D}}_{k},\mathcal{O}_{k+1 \to k})\big)
\big\rVert_{1},
\end{equation}
which reduces flicker and drift in where corresponding. Similarly, surface normal stabilization term:
\begin{equation}
\mathcal{L}_{\mathrm{temp}}^{n}
=\frac{1}{|\mathcal{M}|}\, 
\mathcal{M}\odot\big(1-\mathrm{cos}\langle 
{\mathcal{W}(\hat{\mathbf{N}}_{k},\mathcal{O}_{k \to k+1})},\ \hat{\mathbf{N}}_{k+1}\rangle\big)
+\frac{1}{|\mathcal{M}|}\,
\mathcal{M}\odot\big(1-\mathrm{cos}\langle 
{\mathcal{W}(\hat{\mathbf{N}}_{k+1},\mathcal{O}_{k+1 \to k})},\ \hat{\mathbf{N}}_{k}\rangle\big).
\end{equation}
This term uses a smaller weight than the depth temporal term to suppress random directional jitter without oversmoothing true edges. Finally, the Stage-2 objective is:
\begin{equation}
\mathcal{L}_{\mathrm{stage2}}
=\mathcal{L}_{\mathrm{stage1}}
+\lambda^{d}_{\mathrm{temp}}\mathcal{L}_{\mathrm{temp}}^d
+\lambda^{n}_{\mathrm{temp}}\mathcal{L}_{\mathrm{temp}}^n.
\end{equation}

\begin{table}
\renewcommand{\arraystretch}{1.3}
\setlength{\tabcolsep}{3pt}
\caption{\textbf{Quantitative comparison} for surface normal estimation on THuman2.1 and Hi4D dataset. Note that the parameter size of Sapiens-0.3B is equivalent to that of large models of ViT-based methods.}
\label{tab:static_normal}
\centering
\resizebox{\linewidth}{!}{
\begin{tabular}{lcccccccccc}
\toprule
\multirow{3}{*}{Methods} & \multicolumn{5}{c}{THuman2.1} & \multicolumn{5}{c}{Hi4D}  \\
\cmidrule(lr){2-6} \cmidrule(lr){7-11}
& \multicolumn{2}{c}{Angular Error ($^\circ$) $\downarrow$} & \multicolumn{3}{c}{$\%$ Within $t^\circ$ $\uparrow$} & \multicolumn{2}{c}{Angular Error ($^\circ$) $\downarrow$} & \multicolumn{3}{c}{$\%$ Within $t^\circ$ $\uparrow$} \\
\cmidrule(lr){2-3} \cmidrule(lr){4-6} \cmidrule(lr){7-8} \cmidrule(lr){9-11}
& \multicolumn{1}{c}{Mean} & \multicolumn{1}{c}{Median} & \multicolumn{1}{c}{11.25$^\circ$} & \multicolumn{1}{c}{22.5$^\circ$} & \multicolumn{1}{c}{30$^\circ$} & \multicolumn{1}{c}{Mean} & \multicolumn{1}{c}{Median} & \multicolumn{1}{c}{11.25$^\circ$} & \multicolumn{1}{c}{22.5$^\circ$} & \multicolumn{1}{c}{30$^\circ$} \\
\cmidrule(r){1-1} \cmidrule(lr){2-3} \cmidrule(lr){4-6} \cmidrule(lr){7-8} \cmidrule(l){9-11}
MoGe2-B & 20.31 & 17.94  & 27.04 & 64.96 & 81.30 & \underline{19.29} & \underline{15.52}  & \underline{33.52} & \underline{72.03} & \underline{85.31} \\
DAViD-B & \underline{19.85} & \underline{16.89}  & \underline{31.38} & \underline{67.40} & \underline{81.56} & 20.64 & 16.10  & 32.14 & 69.69 & 82.70 \\
\cellcolor{mypink}Ours-B & \cellcolor{mypink}\textbf{17.89} & \cellcolor{mypink}\textbf{15.56}  & \cellcolor{mypink}\textbf{32.98} & \cellcolor{mypink}\textbf{73.69} & \cellcolor{mypink}\textbf{87.15} & \cellcolor{mypink}\textbf{16.08} & \cellcolor{mypink}\textbf{12.03} & \cellcolor{mypink}\textbf{47.76} & \cellcolor{mypink}\textbf{81.49} & \cellcolor{mypink}\textbf{89.98} \\ 
\cmidrule(r){1-1} \cmidrule(lr){2-3} \cmidrule(lr){4-6} \cmidrule(lr){7-8} \cmidrule(l){9-11}
MoGe2-L & 18.21 & 16.00  & 31.95 & 72.01 & 86.41 & \underline{17.26} & \underline{13.60}  & \underline{40.40} & \underline{78.61} & \underline{88.92} \\
DAViD-L & 19.59 & 16.64  & 30.02 & 68.18 & 82.09 & 20.74 & 16.11  & 31.94 & 69.42 & 82.55 \\
Sapiens-0.3B & \textbf{14.34} & \textbf{11.84}  & \textbf{49.60} & \textbf{83.79} & \textbf{92.07} & 20.01 & 15.42  & 34.41 & 71.58 & 83.90 \\
\cellcolor{mypink}Ours-L & \cellcolor{mypink}\underline{16.00} & \cellcolor{mypink}\underline{13.51}  & \cellcolor{mypink}\underline{41.00} & \cellcolor{mypink}\underline{79.79} & \cellcolor{mypink}\underline{90.04} & \cellcolor{mypink}\textbf{15.00} & \cellcolor{mypink}\textbf{10.84}  & \cellcolor{mypink}\textbf{53.56} & \cellcolor{mypink}\textbf{84.27} & \cellcolor{mypink}\textbf{91.15} \\ 
\cmidrule(r){1-1} \cmidrule(lr){2-3} \cmidrule(lr){4-6} \cmidrule(lr){7-8} \cmidrule(l){9-11}
\color[HTML]{C0C0C0}Sapiens-0.6B & \color[HTML]{C0C0C0}14.34 & \color[HTML]{C0C0C0}11.92  & \color[HTML]{C0C0C0}49.19 & \color[HTML]{C0C0C0}83.82 & \color[HTML]{C0C0C0}92.22 & \color[HTML]{C0C0C0}17.87 & \color[HTML]{C0C0C0}13.50  & \color[HTML]{C0C0C0}41.43 & \color[HTML]{C0C0C0}77.79 & \color[HTML]{C0C0C0}87.79 \\
\color[HTML]{C0C0C0}Sapiens-1B & \color[HTML]{C0C0C0}13.36 & \color[HTML]{C0C0C0}10.91  & \color[HTML]{C0C0C0}54.06 & \color[HTML]{C0C0C0}86.31 & \color[HTML]{C0C0C0}93.38 & \color[HTML]{C0C0C0}15.50 & \color[HTML]{C0C0C0}10.96  & \color[HTML]{C0C0C0}52.93 & \color[HTML]{C0C0C0}83.74 & \color[HTML]{C0C0C0}90.66 \\
\color[HTML]{C0C0C0}Sapiens-2B & \color[HTML]{C0C0C0}13.13 & \color[HTML]{C0C0C0}10.66  & \color[HTML]{C0C0C0}55.38 & \color[HTML]{C0C0C0}86.81 & \color[HTML]{C0C0C0}93.57 & \color[HTML]{C0C0C0}15.58 & \color[HTML]{C0C0C0}11.05  & \color[HTML]{C0C0C0}52.47 & \color[HTML]{C0C0C0}84.02 & \color[HTML]{C0C0C0}90.79 \\ 
\bottomrule
\end{tabular}
}
\end{table}

\section{Experiments}
\subsection{Implementation details}
As described in \Cref{sec:3.3}, we train our model using both static and dynamic data. The static data consists of 2M samples from our synthetic dataset and 300K samples from the SynthHuman dataset \citep{saleh2025david}, while the dynamic data uses 4M samples from our synthetic dataset. We adopt the latest DINOv3 \citep{simeoni2025dinov3} as the pretrained weights. For the static image model, we train a ViT-L with a batch size of 128 for 50K steps, which takes about 2.5 days. For the dynamic video model, we use a batch size of 8 with 32 frames and train for 35K steps, which requires about 1.5 days. The detailed hyperparameters for both training stages are provided in the \Cref{sec:imp}.

\subsection{Evaluation protocol}
\begin{table}
\renewcommand{\arraystretch}{1.3}
\setlength{\tabcolsep}{2pt}
\caption{Comparison on the P3M-500-NP, P3M-500-P and PPM-100 benchmarks.}
\centering
\label{tab:mask}
\begin{tabular}{lcccccccc}
\toprule
\multirow{2}{*}{Method} & \multicolumn{3}{c}{P3M-500-NP} & \multicolumn{3}{c}{P3M-500-P} & \multicolumn{2}{c}{PPM-100} \\
\cmidrule(lr){2-4} \cmidrule(lr){5-7} \cmidrule(l){8-9}
 & SAD $\downarrow$ & SAD-T $\downarrow$ & Conn $\downarrow$
 & SAD $\downarrow$ & SAD-T $\downarrow$ & Conn $\downarrow$
 & SAD $\downarrow$ & Conn $\downarrow$ \\
\cmidrule(r){1-1} \cmidrule(lr){2-4} \cmidrule(lr){5-7} \cmidrule(l){8-9}
Zhong et al. & \textbf{10.60} & \textbf{6.83} & \textbf{9.77} & 10.04 & \textbf{6.44} & \textbf{9.41} & 90.28  & 84.09  \\
BGMv2 & 15.66 & 7.72  & 14.65 & 13.90 & 7.23 & 13.13 & 159.44 & 149.79 \\
P3M-Net & 11.23 & 7.65 & 12.51 & \textbf{8.73}  & 6.89 & 13.88 & 142.74 & 139.89 \\
MODNet & 20.20 & 12.48 & 18.41 & 30.08 & 12.22 & 28.61 & 104.35 & 96.45  \\
DAViD & 14.83 & 10.23 & 14.76 & 12.65 & 9.19 & 12.47 & 78.17 & 74.72 \\
\cmidrule(r){1-1} \cmidrule(lr){2-4} \cmidrule(lr){5-7} \cmidrule(l){8-9}
\cellcolor{mypink}Ours & \cellcolor{mypink}13.12 & \cellcolor{mypink}11.88 & \cellcolor{mypink}12.72 & \cellcolor{mypink}11.63 & \cellcolor{mypink}9.95 & \cellcolor{mypink}11.51 & \cellcolor{mypink}\textbf{70.71} & \cellcolor{mypink}\textbf{68.32} \\
\bottomrule
\end{tabular}
\end{table}

\textbf{Evaluation Datasets.} We evaluate our method on two challenging real-world datasets, THuman2.1 \citep{Thuman} and Hi4D \citep{yin2023hi4d}, for validating depth estimation and surface normal estimation. Following the evaluation protocol in Sapiens \citep{khirodkar2024sapiens}, we construct three subsets on THuman2.1, including face, upper-body, and full-body. Unlike prior works that mainly relied on THuman2.0 with only 500 models and 1,500 images, we adopt the latest THuman2.1 dataset, which contains 2,445 models. Based on these models, we synthesize 7,335 images, resulting in a dataset with a significantly larger scale. For Hi4D, we select sequences from subjects 28, 32, and 37 captured by camera 4, covering 6 different subjects and yielding 1,195 multi-person real images. For image evaluation, we employ both THuman2.1 and Hi4D to assess depth and surface normal estimation under static poses. For video evaluation, we utilize Hi4D, which also provides temporally continuous dynamic sequences, enabling us to further examine the adaptability and generalization of our method in dynamic scenarios. \\\
\textbf{Evaluation Metric.} Following previous work \citep{khirodkar2024sapiens}, to evaluate image depth estimation, we report the mean absolute value of the relative depth (AbsRel) and the root mean square error (RMSE). To evaluate image surface normal estimation, we use the standard metrics of mean and median angular error, as well as the percentage of pixels within $t^\circ$ error for $t \in \{11.25,22.5,30\}$. For video depth and surface normal estimation, we further consider temporal consistency across frames. We employ optical flow-based metrics computed using RAFT \citep{teed2020raft}. We report the optical flow-based warping metric (OPW) \citep{wang2022less}, which measures the discrepancy between consecutive frames after warping. For depth frames, we report the flow-based temporal consistency error (TC-RMSE), which measures the stability of depth predictions across time. For normal frames, we report the flow-based angular error (TC-Mean), which evaluates the temporal consistency of surface normals. However, it should be noted that TC-Mean may be inaccurate if the predicted surface normals are globally biased or too smooth, especially when only evaluating the foreground. Thus, we introduce a new temporal consistency metric for surface normals. Based on the flow-based angular error, we compute the ground truth angular error and compare it with the predicted angular error using the absolute difference (TC-Abs). This metric reflects the discrepancy between predicted and ground truth temporal changes in surface orientation. Unlike purely flow-warped metrics, it can partly mitigate the influence of flow inaccuracies and place more emphasis on whether the temporal variations in predictions follow the ground truth.

\begin{table}
\renewcommand{\arraystretch}{1.2}
\setlength{\tabcolsep}{3pt}
\caption{\textbf{Qualitative comparison} for video depth and surface normal estimation on Hi4D.}
\label{tab:dynamic}
\centering
\begin{tabular}{lccccc}
\toprule
\multirow{2}{*}{Methods} & \multicolumn{2}{c}{Depth} & \multicolumn{3}{c}{Normal} \\
\cmidrule(lr){2-3} \cmidrule(lr){4-6}
 & OPW$\downarrow$ & TC-RMSE$\downarrow$ & OPW$\downarrow$ & TC-Mean$\downarrow$ & TC-Abs$\downarrow$\\
\cmidrule(r){1-1} \cmidrule(lr){2-3} \cmidrule(l){4-6}
MoGe2-B      & 0.0176 & 0.0283 & 0.0362 & 4.26 & 0.162 \\
MoGe2-L      & 0.0176 & 0.0288 & 0.0363 & 4.27 & 0.146 \\
DAViD-B      & 0.0176 & 0.0283 & 0.0423 & 4.92 & 0.170 \\
DAViD-L      & 0.0176 & 0.0288 & 0.0423 & 4.93 & 0.170 \\
Sapiens-0.3B & 0.0145 & 0.0226 & 0.0594 & 6.91 & 0.164 \\
Sapiens-0.6B & 0.0165 & 0.0266 & 0.0486 & 5.64 & 0.147 \\
Sapiens-1B   & 0.0141 & 0.0240 & 0.0452 & 5.26 & 0.147 \\
Sapiens-2B   & 0.0122 & 0.0221 & 0.0421 & 4.89 & 0.149 \\
NormalCrafter& -      & -      & \underline{0.0277} & \underline{3.20} & 0.143 \\
DepthCrafter & 0.0111 & 0.0304 & - & - & - \\
VDA-B        & 0.0111 & 0.0304 & - & - & - \\
VDA-L        & 0.0102 & 0.0300 & - & -  & - \\
\cmidrule(r){1-1} \cmidrule(lr){2-3} \cmidrule(l){4-6}
\cellcolor{mypink}Ours-B &\cellcolor{mypink}\underline{0.0072}        &\cellcolor{mypink}\underline{0.0189} &\cellcolor{mypink}0.0280 &\cellcolor{mypink}3.27 & \cellcolor{mypink}\underline{0.140}\\
\cellcolor{mypink}Ours-L & \cellcolor{mypink}\textbf{0.0070} & \cellcolor{mypink}\textbf{0.0166} & \cellcolor{mypink}\textbf{0.0261} & \cellcolor{mypink}\textbf{3.04} & \cellcolor{mypink}\textbf{0.133}\\
\bottomrule
\end{tabular}
\end{table}

\begin{figure*}
    \centering
    \includegraphics[width=0.9275\linewidth]{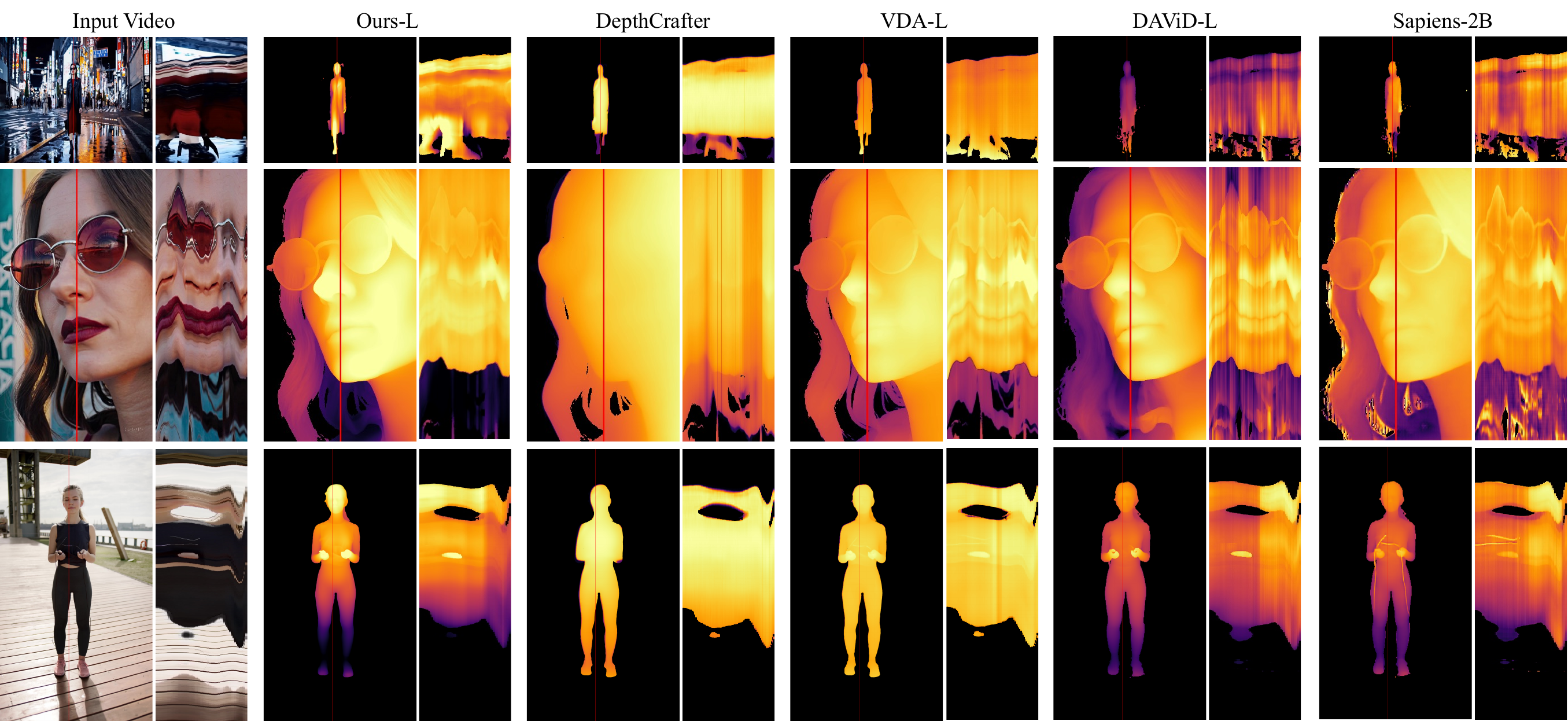}
    \caption{\textbf{Qualitative comparison} on video depth estimation. For better visualization, we also show the time slice on the red lines of each video on their right side.}
    \label{fig:dynamic_depth}
\end{figure*}

\begin{figure*}
    \centering
    \includegraphics[width=0.9275\linewidth]{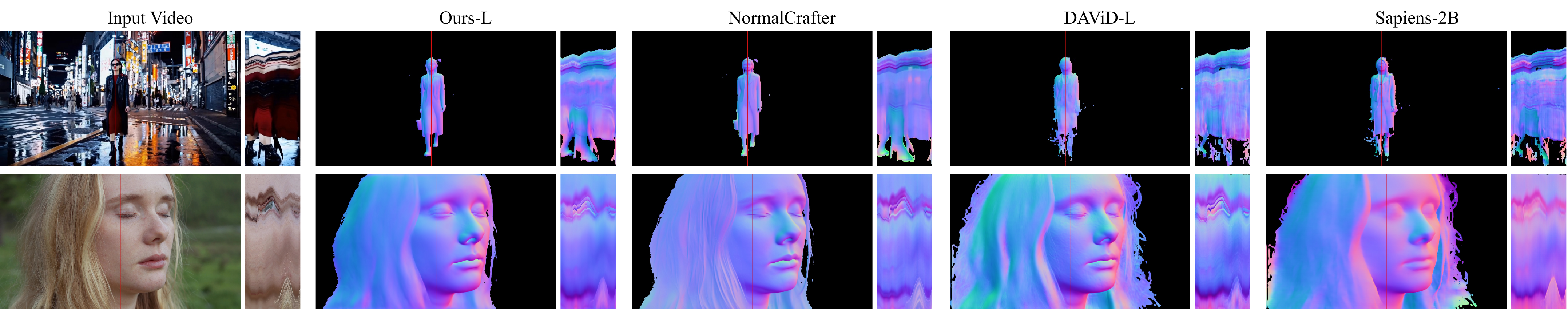}
    \caption{\textbf{Qualitative comparison} on video surface normal estimation. For better visualization, we also show the time slice on the red lines of each video on their right side.}
    \label{fig:dynamic_normal}
\end{figure*}

\begin{table}
\renewcommand{\arraystretch}{1.3}
\setlength{\tabcolsep}{3pt}
\caption{\textbf{Ablation} on Hi4D dataset.}
\centering
\label{tab:abl}
\begin{tabular}{lccccccc}
\toprule
\multirow{2}{*}{Methods} & \multicolumn{2}{c}{Depth} & \multicolumn{5}{c}{Normal}                    \\
\cmidrule(lr){2-3} \cmidrule(lr){4-8} & RMSE $\downarrow$ & AbsRel $\downarrow$ & Mean $\downarrow$ & Median $\downarrow$ & 11.25$^\circ$ $\uparrow$ & 22.5$^\circ$ $\uparrow$ & 30$^\circ$ $\uparrow$\\ 
\cmidrule(r){1-1} \cmidrule(lr){2-3} \cmidrule(l){4-8}
Baseline & 0.0964 & 0.0279 & 20.51 & 16.00 & 32.22 & 70.12 & 82.74 \\
w/ CSE   & 0.0932 & 0.0274 & 17.97 & 14.33 & 40.57 & 76.98 & 88.00 \\
w/ CWA   & 0.0944 & 0.0271 & 18.32 & 15.82 & 42.31 & 77.43 & 88.56 \\ 
Full     & 0.0928 & 0.0277 & 16.08 & 12.03 & 47.76 & 81.49 & 89.98 \\ 
\bottomrule
\end{tabular}
\end{table}

\subsection{Comparison to the state-of-the-art}
For static depth estimation, we evaluate several SOTA models, including general-purpose approaches (the Depth Anything family \citep{yang2024depth,yang2024depthv2} and the Moge family \citep{wang2025moge, wang2025moge2}) as well as human-centric methods (Sapiens \cite{khirodkar2024sapiens} and DAViD \citep{saleh2025david}). As shown in \Cref{tab:static_depth}, both variants of our model outperform these baselines on both datasets. Notably, our Large model achieves comparable or even superior accuracy on Hi4D static depth compared to the larger Sapiens-0.6B/1B/2B, highlighting its parameter efficiency and strong cross-dataset generalization. For static surface normal estimation, the results in \Cref{tab:static_normal} show that Sapiens performs particularly well on the THuman2.1 dataset, likely due to its similarity to RenderPeople, which was used during fine-tuning. On the Hi4D dataset, however, our Large model even surpasses Sapiens-2B. For soft foreground segmentation, we follow the experimental setting of DAViD to compare Zhong et al. \citep{zhong2024lightweight}, BGMv2 \citep{BGMv2}, P3M-Net \citep{rethink_p3m}, and MODNet \citep{ke2022modnet}, and we quantitatively evaluate our segmentation head on the P3M-500-NP, P3M-500-P, and PPM-100 benchmark datasets. As shown in \Cref{tab:mask}, our method shows competitive results on both P3M validation sets and gives clear gains over DAViD and the other baselines on PPM-100, where our approach decreases SAD from 78.17 to 70.71 and Conn from 74.72 to 68.32. Since there is currently no released video human-centric model for depth or surface normal estimation, we compare against SOTA models designed for general scene videos, such as NormalCrafter \citep{bin2025normalcrafter}, DepthCrafter \citep{hu2025depthcrafter}, and VDA \citep{chen2025video}. As shown in \Cref{tab:dynamic}, our model demonstrates superior performance in scenes containing humans. \Cref{fig:static} demonstrates the robustness of our method when tested on person-centric images, covering normal images, shadows, lighting changes, and multi-person scenes. \Cref{fig:dynamic_depth} and \Cref{fig:dynamic_normal} demonstrate the excellent performance of our method on human-centric in-the-wild Internet videos and the temporal consistency, respectively.

\subsection{Ablation Studies}
We mainly conduct the ablation studies on the Hi4D dataset. As shown in \Cref{tab:abl}, we compare our full model against three variants: A) DPT head with an additional CNN branch as the baseline; B) w/ Human CSE prior; C) w/ channel weight adaptation. The results indicate that incorporating human structural priors through CSE encourages the model to capture geometry that aligns with human body shape and articulation, which strengthens local surface details and orientation consistency. On the other hand, CWA emphasizes adaptive feature reweighting across channels, which improves the prediction stability. More detail ablation, please refer to \Cref{sec:abl}.

\section{Conclusion}
This work presented a framework for human-centric dense prediction with temporal consistency. By constructing a synthetic pipeline that produces static frames and dynamic sequences with pixel-accurate annotations, we enabled joint learning of segmentation, depth, and surface normals with both spatial accuracy and stable video performance. Our model achieves strong results on THuman2.1 and Hi4D, and generalizes to in-the-wild videos. The results indicate that large-scale synthetic data, together with temporal supervision and human priors, can be an effective approach for improving human-centric video perception. In future work, we plan to extend the framework to more complex scenes and examine its use in downstream tasks such as human 3D reconstruction.

\bibliographystyle{unsrt}  
\bibliography{references}  

\begin{thebibliography}{10}

\bibitem{xiu2022icon}
Yuliang Xiu, Jinlong Yang, Dimitrios Tzionas, and Michael~J Black.
\newblock Icon: Implicit clothed humans obtained from normals.
\newblock In {\em 2022 IEEE/CVF Conference on Computer Vision and Pattern Recognition (CVPR)}, pages 13286--13296. IEEE, 2022.

\bibitem{weng2022humannerf}
Chung-Yi Weng, Brian Curless, Pratul~P Srinivasan, Jonathan~T Barron, and Ira Kemelmacher-Shlizerman.
\newblock Humannerf: Free-viewpoint rendering of moving people from monocular video.
\newblock In {\em Proceedings of the IEEE/CVF conference on computer vision and pattern Recognition}, pages 16210--16220, 2022.

\bibitem{zhang2023adding}
Lvmin Zhang, Anyi Rao, and Maneesh Agrawala.
\newblock Adding conditional control to text-to-image diffusion models.
\newblock In {\em Proceedings of the IEEE/CVF international conference on computer vision}, pages 3836--3847, 2023.

\bibitem{hu2024animate}
Li~Hu.
\newblock Animate anyone: Consistent and controllable image-to-video synthesis for character animation.
\newblock In {\em Proceedings of the IEEE/CVF Conference on Computer Vision and Pattern Recognition}, pages 8153--8163, 2024.

\bibitem{khirodkar2024sapiens}
Rawal Khirodkar, Timur Bagautdinov, Julieta Martinez, Su~Zhaoen, Austin James, Peter Selednik, Stuart Anderson, and Shunsuke Saito.
\newblock Sapiens: Foundation for human vision models.
\newblock In {\em European Conference on Computer Vision}, pages 206--228. Springer, 2024.

\bibitem{drobyshev2022megaportraits}
Nikita Drobyshev, Jenya Chelishev, Taras Khakhulin, Aleksei Ivakhnenko, Victor Lempitsky, and Egor Zakharov.
\newblock Megaportraits: One-shot megapixel neural head avatars.
\newblock In {\em Proceedings of the 30th ACM International Conference on Multimedia}, pages 2663--2671, 2022.

\bibitem{zhang2019pose2seg}
Song-Hai Zhang, Ruilong Li, Xin Dong, Paul Rosin, Zixi Cai, Xi~Han, Dingcheng Yang, Haozhi Huang, and Shi-Min Hu.
\newblock Pose2seg: Detection free human instance segmentation.
\newblock In {\em Proceedings of the IEEE/CVF conference on computer vision and pattern recognition}, pages 889--898, 2019.

\bibitem{lin2014microsoft}
Tsung-Yi Lin, Michael Maire, Serge Belongie, James Hays, Pietro Perona, Deva Ramanan, Piotr Doll{\'a}r, and C~Lawrence Zitnick.
\newblock Microsoft coco: Common objects in context.
\newblock In {\em European conference on computer vision}, pages 740--755. Springer, 2014.

\bibitem{cao2017realtime}
Zhe Cao, Tomas Simon, Shih-En Wei, and Yaser Sheikh.
\newblock Realtime multi-person 2d pose estimation using part affinity fields.
\newblock In {\em Proceedings of the IEEE conference on computer vision and pattern recognition}, pages 7291--7299, 2017.

\bibitem{saleh2025david}
Fatemeh Saleh, Sadegh Aliakbarian, Charlie Hewitt, Lohit Petikam, Antonio Criminisi, Thomas~J Cashman, Tadas Baltru{\v{s}}aitis, et~al.
\newblock David: Data-efficient and accurate vision models from synthetic data.
\newblock {\em arXiv preprint arXiv:2507.15365}, 2025.

\bibitem{saito2020pifuhd}
Shunsuke Saito, Tomas Simon, Jason Saragih, and Hanbyul Joo.
\newblock Pifuhd: Multi-level pixel-aligned implicit function for high-resolution 3d human digitization.
\newblock In {\em Proceedings of the IEEE/CVF conference on computer vision and pattern recognition}, pages 84--93, 2020.

\bibitem{xiu2023econ}
Yuliang Xiu, Jinlong Yang, Xu~Cao, Dimitrios Tzionas, and Michael~J Black.
\newblock Econ: Explicit clothed humans optimized via normal integration.
\newblock In {\em Proceedings of the IEEE/CVF conference on computer vision and pattern recognition}, pages 512--523, 2023.

\bibitem{chen2025video}
Sili Chen, Hengkai Guo, Shengnan Zhu, Feihu Zhang, Zilong Huang, Jiashi Feng, and Bingyi Kang.
\newblock Video depth anything: Consistent depth estimation for super-long videos.
\newblock In {\em Proceedings of the Computer Vision and Pattern Recognition Conference}, pages 22831--22840, 2025.

\bibitem{AMASS}
Naureen Mahmood, Nima Ghorbani, Nikolaus~F. Troje, Gerard Pons-Moll, and Michael~J. Black.
\newblock {AMASS}: Archive of motion capture as surface shapes.
\newblock In {\em International Conference on Computer Vision}, pages 5442--5451, October 2019.

\bibitem{cse}
Natalia Neverova, David Novotny, Vasil Khalidov, Marc Szafraniec, Patrick Labatut, and Andrea Vedaldi.
\newblock Continuous surface embeddings.
\newblock 2020.

\bibitem{yang2024depth}
Lihe Yang, Bingyi Kang, Zilong Huang, Xiaogang Xu, Jiashi Feng, and Hengshuang Zhao.
\newblock Depth anything: Unleashing the power of large-scale unlabeled data.
\newblock In {\em Proceedings of the IEEE/CVF conference on computer vision and pattern recognition}, pages 10371--10381, 2024.

\bibitem{yang2024depthv2}
Lihe Yang, Bingyi Kang, Zilong Huang, Zhen Zhao, Xiaogang Xu, Jiashi Feng, and Hengshuang Zhao.
\newblock Depth anything v2.
\newblock {\em Advances in Neural Information Processing Systems}, 37:21875--21911, 2024.

\bibitem{simeoni2025dinov3}
Oriane Sim{\'e}oni, Huy~V Vo, Maximilian Seitzer, Federico Baldassarre, Maxime Oquab, Cijo Jose, Vasil Khalidov, Marc Szafraniec, Seungeun Yi, Micha{\"e}l Ramamonjisoa, et~al.
\newblock Dinov3.
\newblock {\em arXiv preprint arXiv:2508.10104}, 2025.

\bibitem{miao2025towards}
Xingyu Miao, Haoran Duan, Quanhao Qian, Jiuniu Wang, Yang Long, Ling Shao, Deli Zhao, Ran Xu, and Gongjie Zhang.
\newblock Towards scalable spatial intelligence via 2d-to-3d data lifting.
\newblock {\em arXiv preprint arXiv:2507.18678}, 2025.

\bibitem{viola2004robust}
Paul Viola and Michael~J Jones.
\newblock Robust real-time face detection.
\newblock {\em International journal of computer vision}, 57(2):137--154, 2004.

\bibitem{andriluka20142d}
Mykhaylo Andriluka, Leonid Pishchulin, Peter Gehler, and Bernt Schiele.
\newblock 2d human pose estimation: New benchmark and state of the art analysis.
\newblock In {\em Proceedings of the IEEE Conference on computer Vision and Pattern Recognition}, pages 3686--3693, 2014.

\bibitem{zhu2012face}
Xiangxin Zhu and Deva Ramanan.
\newblock Face detection, pose estimation, and landmark localization in the wild.
\newblock In {\em 2012 IEEE conference on computer vision and pattern recognition}, pages 2879--2886. IEEE, 2012.

\bibitem{kirillov2023segment}
Alexander Kirillov, Eric Mintun, Nikhila Ravi, Hanzi Mao, Chloe Rolland, Laura Gustafson, Tete Xiao, Spencer Whitehead, Alexander~C Berg, Wan-Yen Lo, et~al.
\newblock Segment anything.
\newblock In {\em Proceedings of the IEEE/CVF international conference on computer vision}, pages 4015--4026, 2023.

\bibitem{wang2025moge}
Ruicheng Wang, Sicheng Xu, Cassie Dai, Jianfeng Xiang, Yu~Deng, Xin Tong, and Jiaolong Yang.
\newblock Moge: Unlocking accurate monocular geometry estimation for open-domain images with optimal training supervision, 2025.

\bibitem{ye2024stablenormal}
Chongjie Ye, Lingteng Qiu, Xiaodong Gu, Qi~Zuo, Yushuang Wu, Zilong Dong, Liefeng Bo, Yuliang Xiu, and Xiaoguang Han.
\newblock Stablenormal: Reducing diffusion variance for stable and sharp normal.
\newblock {\em ACM Transactions on Graphics (TOG)}, 2024.

\bibitem{yin2023hi4d}
Yifei Yin, Chen Guo, Manuel Kaufmann, Juan Zarate, Jie Song, and Otmar Hilliges.
\newblock Hi4d: 4d instance segmentation of close human interaction.
\newblock In {\em Computer Vision and Pattern Recognition (CVPR)}, 2023.

\bibitem{Thuman}
Tao Yu, Zerong Zheng, Kaiwen Guo, Pengpeng Liu, Qionghai Dai, and Yebin Liu.
\newblock Function4d: Real-time human volumetric capture from very sparse consumer rgbd sensors.
\newblock In {\em IEEE Conference on Computer Vision and Pattern Recognition (CVPR2021)}, June 2021.

\bibitem{martinez2024codec}
Julieta Martinez, Emily Kim, Javier Romero, Timur Bagautdinov, Shunsuke Saito, Shoou-I Yu, Stuart Anderson, Michael Zollhöfer, Te-Li Wang, Shaojie Bai, Chenghui Li, Shih-En Wei, Rohan Joshi, Wyatt Borsos, Tomas Simon, Jason Saragih, Paul Theodosis, Alexander Greene, Anjani Josyula, Silvio~Mano Maeta, Andrew~I. Jewett, Simon Venshtain, Christopher Heilman, Yueh-Tung Chen, Sidi Fu, Mohamed Ezzeldin~A. Elshaer, Tingfang Du, Longhua Wu, Shen-Chi Chen, Kai Kang, Michael Wu, Youssef Emad, Steven Longay, Ashley Brewer, Hitesh Shah, James Booth, Taylor Koska, Kayla Haidle, Matt Andromalos, Joanna Hsu, Thomas Dauer, Peter Selednik, Tim Godisart, Scott Ardisson, Matthew Cipperly, Ben Humberston, Lon Farr, Bob Hansen, Peihong Guo, Dave Braun, Steven Krenn, He~Wen, Lucas Evans, Natalia Fadeeva, Matthew Stewart, Gabriel Schwartz, Divam Gupta, Gyeongsik Moon, Kaiwen Guo, Yuan Dong, Yichen Xu, Takaaki Shiratori, Fabian Prada, Bernardo~R. Pires, Bo~Peng, Julia Buffalini, Autumn Trimble, Kevyn McPhail, Melissa Schoeller, and
  Yaser Sheikh.
\newblock {Codec Avatar Studio: Paired Human Captures for Complete, Driveable, and Generalizable Avatars}.
\newblock {\em NeurIPS Track on Datasets and Benchmarks}, 2024.

\bibitem{chen2018cascaded}
Yilun Chen, Zhicheng Wang, Yuxiang Peng, Zhiqiang Zhang, Gang Yu, and Jian Sun.
\newblock Cascaded pyramid network for multi-person pose estimation.
\newblock In {\em Proceedings of the IEEE conference on computer vision and pattern recognition}, pages 7103--7112, 2018.

\bibitem{fang2017rmpe}
Hao-Shu Fang, Shuqin Xie, Yu-Wing Tai, and Cewu Lu.
\newblock Rmpe: Regional multi-person pose estimation.
\newblock In {\em Proceedings of the IEEE international conference on computer vision}, pages 2334--2343, 2017.

\bibitem{huang2017coarse}
Shaoli Huang, Mingming Gong, and Dacheng Tao.
\newblock A coarse-fine network for keypoint localization.
\newblock In {\em Proceedings of the IEEE international conference on computer vision}, pages 3028--3037, 2017.

\bibitem{khirodkar2021multi}
Rawal Khirodkar, Visesh Chari, Amit Agrawal, and Ambrish Tyagi.
\newblock Multi-instance pose networks: Rethinking top-down pose estimation.
\newblock In {\em Proceedings of the IEEE/CVF International conference on computer vision}, pages 3122--3131, 2021.

\bibitem{newell2016stacked}
Alejandro Newell, Kaiyu Yang, and Jia Deng.
\newblock Stacked hourglass networks for human pose estimation.
\newblock In {\em European conference on computer vision}, pages 483--499. Springer, 2016.

\bibitem{papandreou2017towards}
George Papandreou, Tyler Zhu, Nori Kanazawa, Alexander Toshev, Jonathan Tompson, Chris Bregler, and Kevin Murphy.
\newblock Towards accurate multi-person pose estimation in the wild.
\newblock In {\em Proceedings of the IEEE conference on computer vision and pattern recognition}, pages 4903--4911, 2017.

\bibitem{sun2019deep}
Ke~Sun, Bin Xiao, Dong Liu, and Jingdong Wang.
\newblock Deep high-resolution representation learning for human pose estimation.
\newblock In {\em Proceedings of the IEEE/CVF conference on computer vision and pattern recognition}, pages 5693--5703, 2019.

\bibitem{xiao2018simple}
Bin Xiao, Haiping Wu, and Yichen Wei.
\newblock Simple baselines for human pose estimation and tracking.
\newblock In {\em Proceedings of the European conference on computer vision (ECCV)}, pages 466--481, 2018.

\bibitem{xia2017joint}
Fangting Xia, Peng Wang, Xianjie Chen, and Alan~L Yuille.
\newblock Joint multi-person pose estimation and semantic part segmentation.
\newblock In {\em Proceedings of the IEEE conference on computer vision and pattern recognition}, pages 6769--6778, 2017.

\bibitem{xia2016zoom}
Fangting Xia, Peng Wang, Liang-Chieh Chen, and Alan~L Yuille.
\newblock Zoom better to see clearer: Human and object parsing with hierarchical auto-zoom net.
\newblock In {\em European Conference on Computer Vision}, pages 648--663. Springer, 2016.

\bibitem{luo2018macro}
Yawei Luo, Zhedong Zheng, Liang Zheng, Tao Guan, Junqing Yu, and Yi~Yang.
\newblock Macro-micro adversarial network for human parsing.
\newblock In {\em Proceedings of the European conference on computer vision (ECCV)}, pages 418--434, 2018.

\bibitem{gong2018instance}
Ke~Gong, Xiaodan Liang, Yicheng Li, Yimin Chen, Ming Yang, and Liang Lin.
\newblock Instance-level human parsing via part grouping network.
\newblock In {\em Proceedings of the European conference on computer vision (ECCV)}, pages 770--785, 2018.

\bibitem{gong2017look}
Ke~Gong, Xiaodan Liang, Dongyu Zhang, Xiaohui Shen, and Liang Lin.
\newblock Look into person: Self-supervised structure-sensitive learning and a new benchmark for human parsing.
\newblock In {\em Proceedings of the IEEE conference on computer vision and pattern recognition}, pages 932--940, 2017.

\bibitem{fang2018weakly}
Hao-Shu Fang, Guansong Lu, Xiaolin Fang, Jianwen Xie, Yu-Wing Tai, and Cewu Lu.
\newblock Weakly and semi supervised human body part parsing via pose-guided knowledge transfer.
\newblock {\em arXiv preprint arXiv:1805.04310}, 2018.

\bibitem{8765346}
Z.~{Cao}, G.~{Hidalgo Martinez}, T.~{Simon}, S.~{Wei}, and Y.~A. {Sheikh}.
\newblock Openpose: Realtime multi-person 2d pose estimation using part affinity fields.
\newblock {\em IEEE Transactions on Pattern Analysis and Machine Intelligence}, 2019.

\bibitem{bhat2023zoedepth}
Shariq~Farooq Bhat, Reiner Birkl, Diana Wofk, Peter Wonka, and Matthias M{\"u}ller.
\newblock Zoedepth: Zero-shot transfer by combining relative and metric depth.
\newblock {\em arXiv preprint arXiv:2302.12288}, 2023.

\bibitem{yin2023metric3d}
Wei Yin, Chi Zhang, Hao Chen, Zhipeng Cai, Gang Yu, Kaixuan Wang, Xiaozhi Chen, and Chunhua Shen.
\newblock Metric3d: Towards zero-shot metric 3d prediction from a single image.
\newblock In {\em Proceedings of the IEEE/CVF international conference on computer vision}, pages 9043--9053, 2023.

\bibitem{jafarian2021learning}
Yasamin Jafarian and Hyun~Soo Park.
\newblock Learning high fidelity depths of dressed humans by watching social media dance videos.
\newblock In {\em Proceedings of the IEEE/CVF Conference on Computer Vision and Pattern Recognition}, pages 12753--12762, 2021.

\bibitem{birkl2023midas}
Reiner Birkl, Diana Wofk, and Matthias M{\"u}ller.
\newblock Midas v3. 1--a model zoo for robust monocular relative depth estimation.
\newblock {\em arXiv preprint arXiv:2307.14460}, 2023.

\bibitem{eigen2015predicting}
David Eigen and Rob Fergus.
\newblock Predicting depth, surface normals and semantic labels with a common multi-scale convolutional architecture.
\newblock In {\em Proceedings of the IEEE international conference on computer vision}, pages 2650--2658, 2015.

\bibitem{ladicky2014discriminatively}
L’ubor Ladick{\`y}, Bernhard Zeisl, and Marc Pollefeys.
\newblock Discriminatively trained dense surface normal estimation.
\newblock In {\em European conference on computer vision}, pages 468--484. Springer, 2014.

\bibitem{oquab2023dinov2}
Maxime Oquab, Timoth{\'e}e Darcet, Th{\'e}o Moutakanni, Huy Vo, Marc Szafraniec, Vasil Khalidov, Pierre Fernandez, Daniel Haziza, Francisco Massa, Alaaeldin El-Nouby, et~al.
\newblock Dinov2: Learning robust visual features without supervision.
\newblock {\em arXiv preprint arXiv:2304.07193}, 2023.

\bibitem{ronneberger2015u}
Olaf Ronneberger, Philipp Fischer, and Thomas Brox.
\newblock U-net: Convolutional networks for biomedical image segmentation.
\newblock In {\em International Conference on Medical image computing and computer-assisted intervention}, pages 234--241. Springer, 2015.

\bibitem{chen2018encoder}
Liang-Chieh Chen, Yukun Zhu, George Papandreou, Florian Schroff, and Hartwig Adam.
\newblock Encoder-decoder with atrous separable convolution for semantic image segmentation.
\newblock In {\em Proceedings of the European conference on computer vision (ECCV)}, pages 801--818, 2018.

\bibitem{ranftl2021vision}
Ren{\'e} Ranftl, Alexey Bochkovskiy, and Vladlen Koltun.
\newblock Vision transformers for dense prediction.
\newblock In {\em Proceedings of the IEEE/CVF international conference on computer vision}, pages 12179--12188, 2021.

\bibitem{dosovitskiy2020image}
Alexey Dosovitskiy, Lucas Beyer, Alexander Kolesnikov, Dirk Weissenborn, Xiaohua Zhai, Thomas Unterthiner, Mostafa Dehghani, Matthias Minderer, Georg Heigold, Sylvain Gelly, et~al.
\newblock An image is worth 16x16 words: Transformers for image recognition at scale.
\newblock {\em arXiv preprint arXiv:2010.11929}, 2020.

\bibitem{ke2023repurposing}
Bingxin Ke, Anton Obukhov, Shengyu Huang, Nando Metzger, Rodrigo~Caye Daudt, and Konrad Schindler.
\newblock Repurposing diffusion-based image generators for monocular depth estimation.
\newblock In {\em Proceedings of the IEEE/CVF Conference on Computer Vision and Pattern Recognition (CVPR)}, 2024.

\bibitem{cimpoi14describing}
M.~Cimpoi, S.~Maji, I.~Kokkinos, S.~Mohamed, , and A.~Vedaldi.
\newblock Describing textures in the wild.
\newblock In {\em Proceedings of the {IEEE} Conf. on Computer Vision and Pattern Recognition ({CVPR})}, 2014.

\bibitem{burghouts2009material}
Gertjan~J Burghouts and Jan-Mark Geusebroek.
\newblock Material-specific adaptation of color invariant features.
\newblock {\em Pattern Recognition Letters}, 30(3):306--313, 2009.

\bibitem{guo2023animatediff}
Yuwei Guo, Ceyuan Yang, Anyi Rao, Zhengyang Liang, Yaohui Wang, Yu~Qiao, Maneesh Agrawala, Dahua Lin, and Bo~Dai.
\newblock Animatediff: Animate your personalized text-to-image diffusion models without specific tuning.
\newblock {\em arXiv preprint arXiv:2307.04725}, 2023.

\bibitem{guler2018densepose}
R{\i}za~Alp G{\"u}ler, Natalia Neverova, and Iasonas Kokkinos.
\newblock Densepose: Dense human pose estimation in the wild.
\newblock In {\em Proceedings of the IEEE conference on computer vision and pattern recognition}, pages 7297--7306, 2018.

\bibitem{hu2019revisiting}
Junjie Hu, Mete Ozay, Yan Zhang, and Takayuki Okatani.
\newblock Revisiting single image depth estimation: Toward higher resolution maps with accurate object boundaries.
\newblock In {\em 2019 IEEE winter conference on applications of computer vision (WACV)}, pages 1043--1051. IEEE, 2019.

\bibitem{teed2020raft}
Zachary Teed and Jia Deng.
\newblock Raft: Recurrent all-pairs field transforms for optical flow.
\newblock In {\em European conference on computer vision}, pages 402--419. Springer, 2020.

\bibitem{wang2022less}
Yiran Wang, Zhiyu Pan, Xingyi Li, Zhiguo Cao, Ke~Xian, and Jianming Zhang.
\newblock Less is more: Consistent video depth estimation with masked frames modeling.
\newblock In {\em Proceedings of the 30th ACM International Conference on Multimedia}, pages 6347--6358, 2022.

\bibitem{wang2025moge2}
Ruicheng Wang, Sicheng Xu, Yue Dong, Yu~Deng, Jianfeng Xiang, Zelong Lv, Guangzhong Sun, Xin Tong, and Jiaolong Yang.
\newblock Moge-2: Accurate monocular geometry with metric scale and sharp details, 2025.

\bibitem{zhong2024lightweight}
Yatao Zhong and Ilya Zharkov.
\newblock Lightweight portrait matting via regional attention and refinement.
\newblock In {\em Proceedings of the IEEE/CVF Winter Conference on Applications of Computer Vision}, pages 4158--4167, 2024.

\bibitem{BGMv2}
Shanchuan Lin, Andrey Ryabtsev, Soumyadip Sengupta, Brian Curless, Steve Seitz, and Ira Kemelmacher-Shlizerman.
\newblock Real-time high-resolution background matting.
\newblock {\em arXiv}, pages arXiv--2012, 2020.

\bibitem{rethink_p3m}
Sihan Ma, Jizhizi Li, Jing Zhang, He~Zhang, and Dacheng Tao.
\newblock Rethinking portrait matting with pirvacy preserving.
\newblock {\em International Journal of Computer Vision}, 2023.

\bibitem{ke2022modnet}
Zhanghan Ke, Jiayu Sun, Kaican Li, Qiong Yan, and Rynson~WH Lau.
\newblock Modnet: Real-time trimap-free portrait matting via objective decomposition.
\newblock In {\em Proceedings of the AAAI Conference on Artificial Intelligence}, volume~36, pages 1140--1147, 2022.

\bibitem{bin2025normalcrafter}
Yanrui Bin, Wenbo Hu, Haoyuan Wang, Xinya Chen, and Bing Wang.
\newblock Normalcrafter: Learning temporally consistent normals from video diffusion priors.
\newblock {\em arXiv preprint arXiv:2504.11427}, 2025.

\bibitem{hu2025depthcrafter}
Wenbo Hu, Xiangjun Gao, Xiaoyu Li, Sijie Zhao, Xiaodong Cun, Yong Zhang, Long Quan, and Ying Shan.
\newblock Depthcrafter: Generating consistent long depth sequences for open-world videos.
\newblock In {\em Proceedings of the Computer Vision and Pattern Recognition Conference}, pages 2005--2015, 2025.

\bibitem{varol2017learning}
Gul Varol, Javier Romero, Xavier Martin, Naureen Mahmood, Michael~J Black, Ivan Laptev, and Cordelia Schmid.
\newblock Learning from synthetic humans.
\newblock In {\em Proceedings of the IEEE conference on computer vision and pattern recognition}, pages 109--117, 2017.

\bibitem{ebadi2021peoplesanspeople}
Salehe~Erfanian Ebadi, You-Cyuan Jhang, Alex Zook, Saurav Dhakad, Adam Crespi, Pete Parisi, Steven Borkman, Jonathan Hogins, and Sujoy Ganguly.
\newblock Peoplesanspeople: a synthetic data generator for human-centric computer vision.
\newblock {\em arXiv preprint arXiv:2112.09290}, 2021.

\bibitem{yang2023synbody}
Zhitao Yang, Zhongang Cai, Haiyi Mei, Shuai Liu, Zhaoxi Chen, Weiye Xiao, Yukun Wei, Zhongfei Qing, Chen Wei, Bo~Dai, et~al.
\newblock Synbody: Synthetic dataset with layered human models for 3d human perception and modeling.
\newblock In {\em Proceedings of the IEEE/CVF International Conference on Computer Vision}, pages 20282--20292, 2023.

\bibitem{hewitt2024look}
Charlie Hewitt, Fatemeh Saleh, Sadegh Aliakbarian, Lohit Petikam, Shideh Rezaeifar, Louis Florentin, Zafiirah Hosenie, Thomas~J Cashman, Julien Valentin, Darren Cosker, and Tadas Baltru\v{s}aitis.
\newblock Look ma, no markers: holistic performance capture without the hassle.
\newblock {\em ACM Transactions on Graphics (TOG)}, 43(6), 2024.

\end{thebibliography}

\newpage
\appendix
\section{Appendix}
\subsection{Data synthesis pipeline}
\begin{figure}[!bth]
    \centering
    \includegraphics[width=\linewidth]{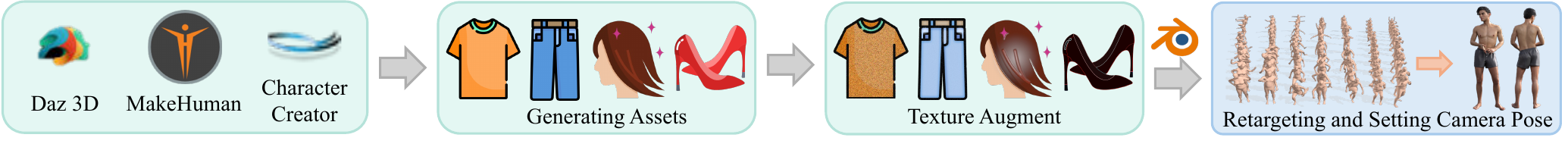}
    \caption{We first generate clothed human models using DAZ 3D, MakeHuman, and Character Creator. Texture augmentations are applied to increase appearance diversity. Each model is then animated by retargeting AMASS motion sequences. Finally, models are placed in Blender with randomized cameras for rendering RGB images together with depth, surface normals, and segmentation masks.}
    \label{fig:data_pipeline}
\end{figure}

The full data synthesis pipeline is illustrated in \Cref{fig:data_pipeline}. The process begins with the construction of clothed human models using character-generation tools such as DAZ 3D, MakeHuman, and Character Creator. These tools provide parametric control over body shape and garment categories, forming the base set of assets used for large-scale identity sampling.

To expand the appearance diversity, we apply texture augmentations on the diffuse maps. These operations include color-based perturbations and material replacement, which allow the same geometry to support a wide range of surface styles. After texture augmentation, each identity is paired with motion data by retargeting AMASS skeletal trajectories. This step assigns realistic human motion while maintaining consistent rigging across different characters.

The animated models are then placed into Blender, where we define camera poses, focal lengths, and tracking behavior. Randomization in these settings increases viewpoint variety in both static and dynamic supervision. During rendering, Blender outputs synchronized RGB images, depth maps, surface normals, and segmentation masks.

\subsection{Discussion}
Existing synthetic data pipelines for human-centric learning tasks primarily focus on static image generation or structural parameter supervision. SURREAL \citep{varol2017learning} combines SMPL models with MoCap sequences to produce synthetic videos with depth, surface normal, and part segmentation annotations. However, it lacks realistic clothing or hair geometry, relying on simplified texture mappings. PeopleSansPeople \citep{ebadi2021peoplesanspeople} leverages Unity to render large-scale, domain-randomized human images, supporting segmentation and keypoint labels but does not generate temporally aligned sequences or pixel-level geometric cues. SynBody \citep{yang2023synbody} substantially improves the scale and diversity of identities and actions using SMPL-XL and layered clothing models, providing multi-view video sequences and mesh-level supervision. However, its released data focuses on RGB and pose annotations, with depth and normal modalities not included in the official release. More recent pipelines such as SynthMoCap \citep{hewitt2024look} emphasize high-fidelity single-frame supervision for dense prediction tasks. They provide detailed annotations like depth, surface normals, and masks, but are limited to frame-level modeling without temporal continuity. Our pipeline is explicitly designed to generate temporally aligned, multi-modal human video sequences, enabling supervision for both per-frame and sequence-level tasks. We synthesize high-quality clothed human characters using commercial modeling tools with randomized sampling over body types, clothing, and textures to construct a large and diverse identity set. Motions from the AMASS dataset are retargeted to these characters and rendered in Blender to produce videos along with per-frame RGB, depth, surface normal, and segmentation maps. Unlike most prior work, our pipeline produces frame-consistent annotations, supporting dense, temporally stable supervision across tasks such as point tracking, normal prediction, and temporally coherent segmentation.

\subsection{Additional qualitative results}
\begin{figure}
    \centering
    \includegraphics[height=\textheight]{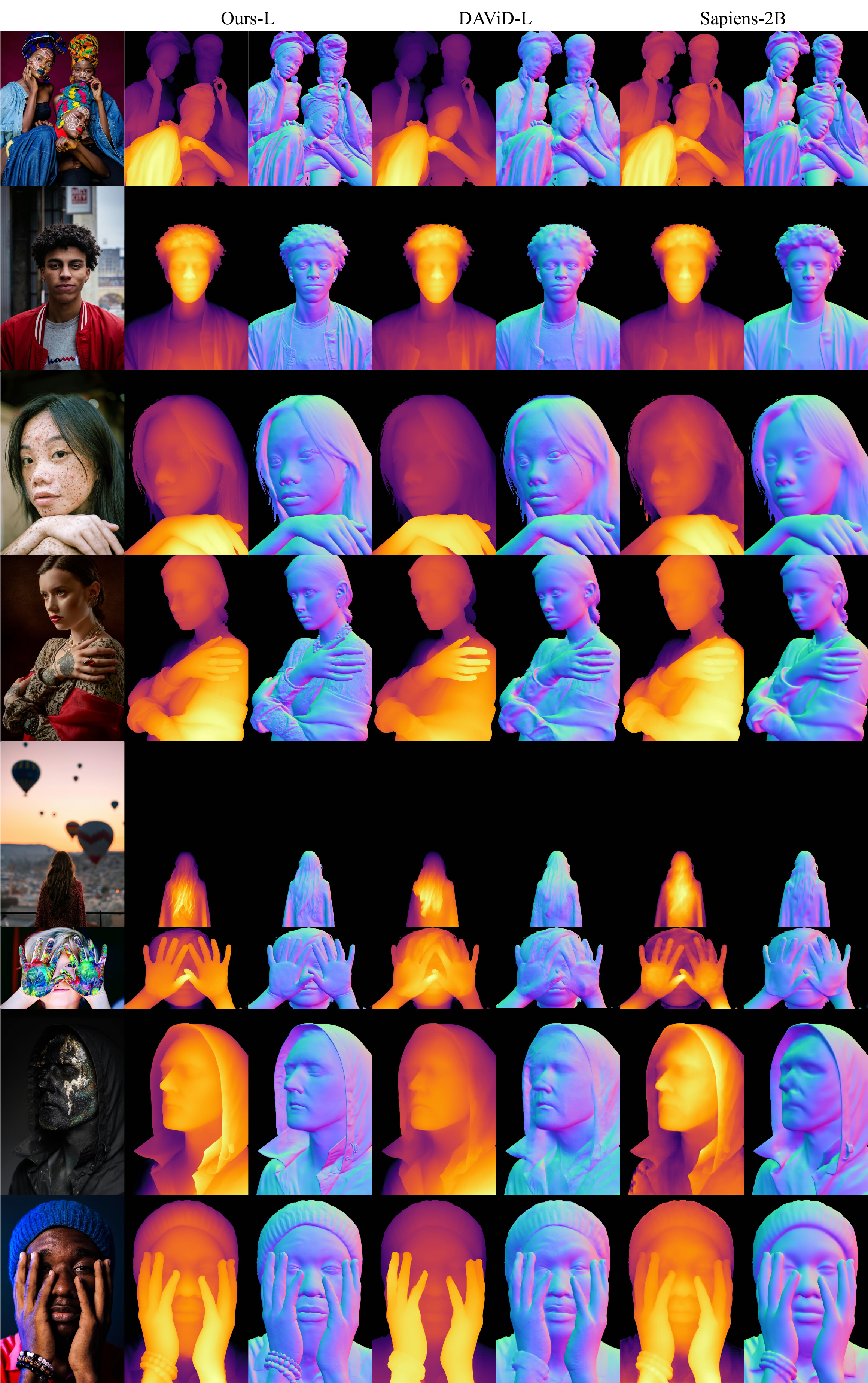}
    \caption{Additional qualitative comparison on challenging images in the wild.}
    \label{fig:supp_qualitative_s_1}
\end{figure}

\begin{figure}
    \centering
    \includegraphics[height=\textheight]{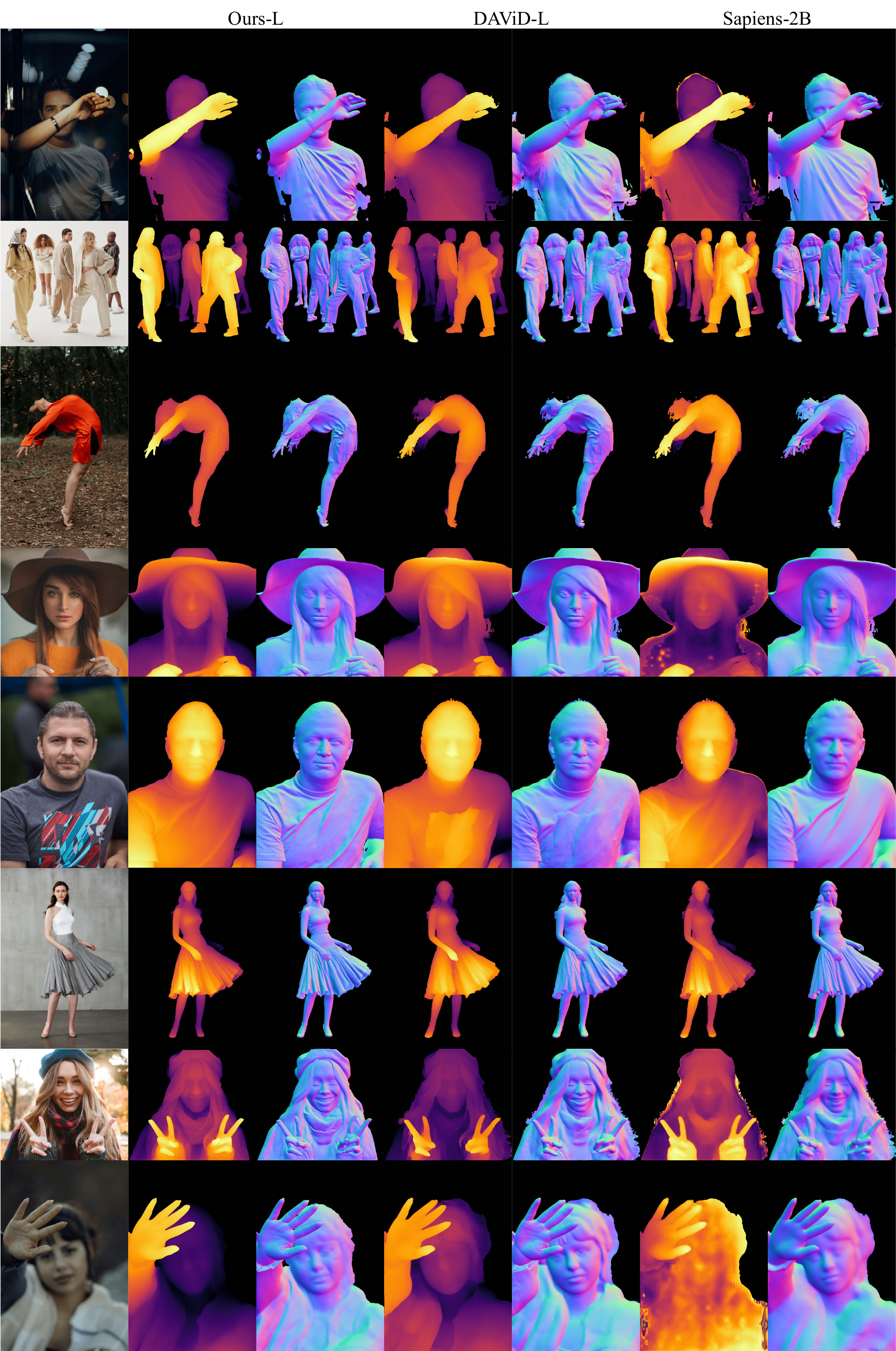}
    \caption{Additional qualitative comparison on challenging images in the wild.}
    \label{fig:supp_qualitative_s_2}
\end{figure}

In \Cref{fig:supp_qualitative_s_1} and \Cref{fig:supp_qualitative_s_2}, we provide additional qualitative results. We also provide some video results in the supplementary materials.

\begin{figure}
    \centering
    \includegraphics[width=\linewidth]{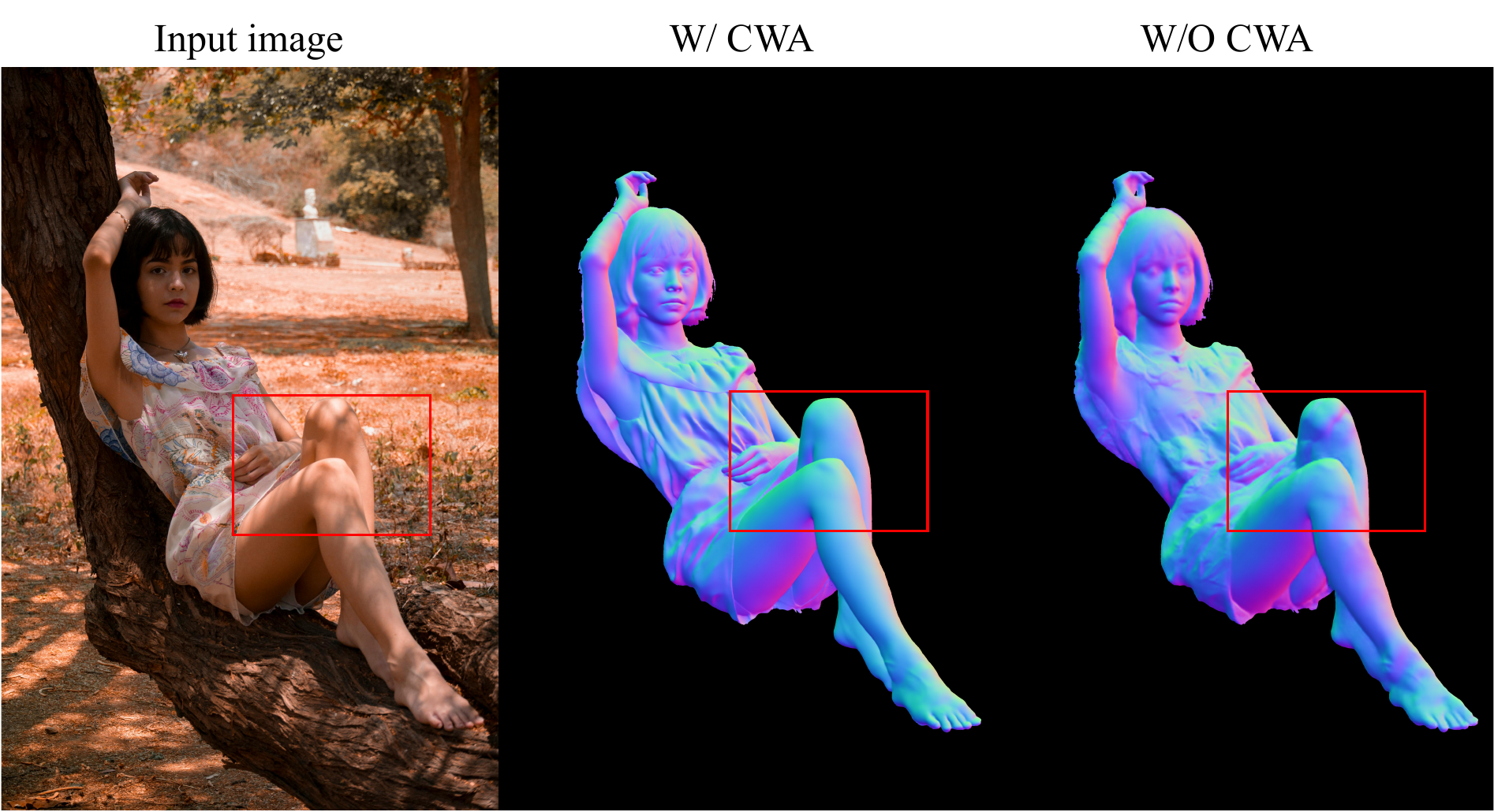}
    \caption{Ablation for Channel Weight Adaptation.}
    \label{fig:ab1}
\end{figure}

\subsection{Additional ablation study}
\label{sec:abl}
\textbf{Investigate channel weight adaptation.} The motivation for introducing a CNN branch in the dual-branch structure is to compensate for the limited capability of the Transformer backbone in modeling local textures, similar to the design adopted in the DAViD \citep{saleh2025david}. CNNs are effective at capturing local patterns and edge details, which helps refine the quality of depth or normal predictions. In our experiments, we observed that adding the CNN branch improves surface continuity and local smoothness. However, this advantage also comes with a drawback, since CNNs tend to capture redundant texture signals that are not related to geometry, such as shadows, clothing patterns, or tattoos. When such signals are fused into the prediction, they interfere with the recovery of the underlying geometry and may produce artifacts or instability in challenging scenarios.

To address this issue, we introduce the CWA module into the CNN branch. CWA adaptively adjusts the channel-wise feature weights, suppressing those that contribute little or negatively to geometry recovery while emphasizing features that are strongly correlated with shape. In practice, CWA acts as a dynamic filter placed between the CNN branch and the final prediction, enabling the model to better distinguish between texture information and geometric cues. Comparative results in \Cref{fig:ab1} show that incorporating CWA effectively reduces artifacts in local regions, especially in cases with complex lighting or decorative textures, and leads to more stable and consistent predictions.

\begin{table}[!bth]
\renewcommand{\arraystretch}{1.3}
\setlength{\tabcolsep}{3pt}
\caption{Ablation for DINOv2 and DINOv3 on Thuman2.1 and Hi4D datasets using the depth task.}
\centering
\label{tab:dino_comp}
\resizebox{0.7\linewidth}{!}{
\begin{tabular}{lcccccccc}
\toprule
\multirow{2}{*}{Methods} & \multicolumn{2}{c}{TH2.1-Face} & \multicolumn{2}{c}{TH2.1-UpperBody} & \multicolumn{2}{c}{TH2.1-FullBody} & \multicolumn{2}{c}{Hi4D} \\
\cmidrule(lr){2-3} \cmidrule(lr){4-5} \cmidrule(lr){6-7} \cmidrule(l){8-9}
& RMSE$\downarrow$ & AbsRel$\downarrow$ & RMSE$\downarrow$ & AbsRel$\downarrow$ & RMSE$\downarrow$ & AbsRel$\downarrow$ & RMSE$\downarrow$ & AbsRel$\downarrow$ \\
\cmidrule(r){1-1} \cmidrule(lr){2-3} \cmidrule(lr){4-5} \cmidrule(lr){6-7} \cmidrule(l){8-9}
DINOv2-B & 0.0207 & 0.0105 & 0.0251 & 0.0116 & 0.0321 & 0.0136 & 0.0862 & 0.0228 \\
DINOv3-B & 0.0193 & 0.0108 & 0.0234 & 0.0116 & 0.0302 & 0.0135 & 0.0871 & 0.0212 \\
DINOv2-L & 0.0167 & 0.0088 & 0.0211 & 0.0102 & 0.0293 & 0.0123 & 0.0771 & 0.0193 \\
DINOv3-L & 0.0158 & 0.0085 & 0.0198 & 0.0098 & 0.0243 & 0.0111 & 0.0768 & 0.0195 \\
\bottomrule
\end{tabular}
}
\end{table}

\textbf{Investigate DINOv2 and DINOv3.} To ensure a fair comparison, we are using $518\times518$ input resolution for DINOv2 \citep{oquab2023dinov2} and $592\times592$ input resolution for DINOv3. As shown in \Cref{tab:dino_comp}, DINOv3 \citep{simeoni2025dinov3} achieves consistent improvements over DINOv2 across multiple benchmarks, particularly on Thuman-FullBody and Hi4D, where the results are more stable. While these gains partially benefit from the stronger encoder, it is important to highlight that the overall performance improvements of our approach do not stem solely from replacing the backbone. Instead, they result from the joint contribution of our carefully designed model and the curated data used for training. This synergy allows the network to better handle the challenges of real-world scenarios, leading to more reliable geometry recovery and stronger generalization across datasets.

\textbf{Investigate human prior.} We investigate human priors on the depth task. As shown in Table \Cref{tab:abl_human_depth}, adding a Fourier  UV map improves over the baseline, suggesting that canonical UV coordinates provide useful geometric cues for depth estimation. While our human prior achieves the more excellent results on both Thuman and Hi4D, reducing both RMSE and AbsRel. These results highlight that human-structure-aware priors enable more accurate and stable depth predictions compared to purely positional encodings.
\begin{table}[!bth]
\renewcommand{\arraystretch}{1.3}
\setlength{\tabcolsep}{3pt}
\caption{Ablation on human priors using the depth task with DINOv3-B.}
\centering
\label{tab:abl_human_depth}
\resizebox{0.45\linewidth}{!}{
\begin{tabular}{lcccc}
\toprule
\multirow{2}{*}{Methods} & \multicolumn{2}{c}{TH2.1} & \multicolumn{2}{c}{Hi4D} \\
\cmidrule(lr){2-3} \cmidrule(l){4-5}
& RMSE$\downarrow$ & AbsRel$\downarrow$ & RMSE$\downarrow$ & AbsRel$\downarrow$ \\
\midrule
Baseline     & 0.0223 & 0.0119 & 0.0964 & 0.0279 \\
+ Fourier UV & 0.0192 & 0.0110 & 0.0922 & 0.0273 \\
+ CSE           & 0.0189 & 0.0108 & 0.0912 & 0.0271 \\
\bottomrule
\end{tabular}
}
\end{table}

\textbf{Investigate human prior fusion strategies.} We compare two strategies for integrating human priors, concatenation (cat) and addition (add). As shown in \Cref{tab:abl_human_integration}, both strategies improve depth and normal estimation, but addition consistently achieves better results. Specifically, add reduces both RMSE and AbsRel while also lowering the mean and median angular error for surface normals. The improvement can be attributed to the fact that addition enforces direct feature alignment between the prior and the learned representations, whereas concatenation requires the network to learn how to fuse heterogeneous features. This suggests that additive integration provides a more effective way to inject human-structure priors, yielding more stable and geometry-aware predictions.

\begin{table}[!bth]
\renewcommand{\arraystretch}{1.3}
\setlength{\tabcolsep}{3pt}
\caption{Ablation on different integration strategies of human priors on Hi4D with DINOv3-B.}
\centering
\label{tab:abl_human_integration}
\resizebox{0.65\linewidth}{!}{
\begin{tabular}{lcccccccc}
\toprule
\multirow{2}{*}{Methods} & \multicolumn{2}{c}{Depth} & \multicolumn{5}{c}{Normal} \\
\cmidrule(lr){2-3} \cmidrule(l){4-8}
& RMSE$\downarrow$ & AbsRel$\downarrow$ & Mean$\downarrow$ & Median$\downarrow$ & 11.25$^\circ$$\uparrow$ & 22.5$^\circ$$\uparrow$ & 30$^\circ$$\uparrow$ \\
\cmidrule(lr){1-1} \cmidrule(lr){2-3} \cmidrule(l){4-8}
Cat & 0.0955 & 0.0284 & 16.96 & 14.23 & 38.20 & 76.13 & 87.86 \\
Add & 0.0933 & 0.0280 & 16.41 & 13.66 & 40.83 & 77.57 & 88.67 \\
\bottomrule
\end{tabular}
}
\end{table}

\textbf{Investigate the impact of loss weight on multi-task.} We further investigate the effect of balancing depth and normal losses by varying the weights $\lambda_{d}$ and $\lambda_{n}$. As shown in \Cref{tab:abl_diff_weights}, setting equal weights ($\lambda_{d}$=1, $\lambda_{n}$=1) gives the weakest performance, suggesting that treating the two tasks uniformly introduces conflicts in optimization. Reducing the normal weight to $\lambda_{n}$=0.5 keeps the depth metrics almost unchanged but leads to a noticeable drop in normal estimation accuracy, indicating that the depth signal dominates training. Increasing the normal weight ($\lambda_{d}$=0.5, $\lambda_{n}$=1) slightly improves surface normals compared to the 1:1 setting but does not yield significant gains. The best results are obtained when $\lambda_{d}$=1 and $\lambda_{n}$=0.1, where both depth and normal predictions improve. This demonstrates that depth supervision should remain the primary training signal, while a lightly weighted normal loss provides complementary regularization without overwhelming the optimization.
\begin{table}[!bth]
\renewcommand{\arraystretch}{1.3}
\setlength{\tabcolsep}{3pt}
\centering
\caption{Ablation for depth and normal loss weights on Hi4D with DINOv3-B.}
\label{tab:abl_diff_weights}
\resizebox{0.65\linewidth}{!}{
\begin{tabular}{lcccccccc}
\toprule
\multirow{2}{*}{Methods} & \multicolumn{2}{c}{Depth} & \multicolumn{5}{c}{Normal} \\
\cmidrule(lr){2-3} \cmidrule(l){4-8}
& RMSE$\downarrow$ & AbsRel$\downarrow$ & Mean$\downarrow$ & Median$\downarrow$ & 11.25$^\circ$$\uparrow$ & 22.5$^\circ$$\uparrow$ & 30$^\circ$$\uparrow$ \\
\midrule
$\lambda_{d}$=1, $\lambda_{n}$=1   & 0.0955 & 0.0284 & 16.96 & 14.23 & 38.20 & 76.13 & 87.86 \\
$\lambda_{d}$=1, $\lambda_{n}$=0.5 & 0.0932 & 0.0281 & 17.25 & 14.50 & 37.60 & 75.98 & 87.40 \\
$\lambda_{d}$=0.5, $\lambda_{n}$=1 & 0.0947 & 0.0283 & 16.70 & 14.05 & 38.85 & 76.80 & 87.95 \\
$\lambda_{d}$=1, $\lambda_{n}$=0.1 & 0.0929 & 0.0279 & 16.61 & 13.76 & 40.03 & 77.37 & 88.55 \\
\bottomrule
\end{tabular}
}
\end{table}

\textbf{Investigate the impact of training data size.}
\Cref{tab:abl_data} reports the impact of training data size on Hi4D using ViT-B. We observe a consistent improvement across both depth and normal prediction as the number of training samples increases. For depth estimation, RMSE and AbsRel gradually decrease when scaling from 300K to 2M, showing that additional data helps the model capture finer geometric cues. A similar trend is observed in surface normal prediction, where both Mean and Median angular errors become smaller, while the percentage of pixels within 11.25$^\circ$, 22.5$^\circ$, and 30$^\circ$ steadily increases. These results suggest that enlarging the training set enhances generalization ability and reduces overfitting, even when the backbone is fixed. However, the gain becomes marginal when moving from 600K to 2M, indicating that data scaling alone may saturate and further improvements may require stronger architectures or better data diversity.

\begin{table}[h]
\renewcommand{\arraystretch}{1.3}
\setlength{\tabcolsep}{3pt}
\centering
\caption{Ablation for different data size on Hi4D with DINOv3-B.}
\label{tab:abl_data}
\resizebox{0.75\linewidth}{!}{
\begin{tabular}{lcccccccc}
\toprule
\multirow{2}{*}{DataSize} & \multicolumn{2}{c}{Depth} & \multicolumn{5}{c}{Normal} \\
\cmidrule(lr){2-3} \cmidrule(l){4-8}
& RMSE$\downarrow$ & AbsRel$\downarrow$ & Mean$\downarrow$ & Median$\downarrow$ & 11.25$^\circ$$\uparrow$ & 22.5$^\circ$$\uparrow$ & 30$^\circ$$\uparrow$ \\
\midrule
Our [300K] & 0.0963 & 0.0286 & 17.10 & 14.40 & 38.10 & 76.10 & 87.80 \\
Our [600K] & 0.0954 & 0.0284 & 16.90 & 14.20 & 38.80 & 76.60 & 88.05 \\
Our [2M]   & 0.0943 & 0.0282 & 16.70 & 13.90 & 39.50 & 76.90 & 88.15 \\
SynthHuman & 0.0971 & 0.0287 & 17.25 & 14.55 & 37.80 & 75.90 & 87.40 \\
SynthHuman + Our [300K] & 0.0958 & 0.0285 & 17.00 & 14.32 & 38.45 & 76.30 & 87.95 \\
SynthHuman + Our [600K] & 0.0946 & 0.0283 & 16.72 & 14.05 & 39.30 & 76.95 & 88.25 \\
SynthHuman + Our [2M] & 0.0940 & 0.0281 & 16.58 & 13.70 & 40.10 & 77.20 & 88.30 \\
\bottomrule
\end{tabular}
}
\end{table}

\textbf{Investigate the impact of normal regularization term.} \Cref{tab:abl_normal_loss_term} evaluates the effect of introducing a normal regularization term (NRT) in training. While the depth estimation metrics remain nearly unchanged, we observe a significant improvement in surface normal prediction.  This indicates that the NRT provides strong geometric guidance, making the network more sensitive to local surface orientation without sacrificing depth accuracy. The results highlight that explicit geometric priors can complement photometric supervision and lead to better normal recovery, even under the same backbone capacity.

\begin{table}[!bth]
\renewcommand{\arraystretch}{1.3}
\setlength{\tabcolsep}{3pt}
\centering
\caption{Ablation for normal regularization term on Hi4D with DINOv3-B.}
\label{tab:abl_normal_loss_term}
\resizebox{0.55\linewidth}{!}{
\begin{tabular}{lcccccccc}
\toprule
\multirow{2}{*}{Methods} & \multicolumn{2}{c}{Depth} & \multicolumn{5}{c}{Normal} \\
\cmidrule(lr){2-3} \cmidrule(l){4-8}
& RMSE$\downarrow$ & AbsRel$\downarrow$ & Mean$\downarrow$ & Median$\downarrow$ & 11.25$^\circ$$\uparrow$ & 22.5$^\circ$$\uparrow$ & 30$^\circ$$\uparrow$ \\
\midrule
w/o NRT   & 0.0938 & 0.0282 & 16.85 & 14.12 & 38.10 & 76.05 & 87.70 \\
w/ NRT    & 0.0935 & 0.0281 & 15.40 & 12.95 & 45.25 & 79.85 & 89.45 \\
\bottomrule
\end{tabular}
}
\end{table}

\textbf{Investigate the impact of the flow stable term.} \Cref{tab:abl_dynamic} investigates the role of different temporal losses. Without temporal regularization, both depth and surface normal predictions are unstable, yielding results on par with Sapiens-1B. Introducing the GMT loss improves depth consistency, with OPW and TC-RMSE reduced compared to the variant without temporal loss. However, the overall performance remains worse than VDA-B, and the normal metrics show almost no improvement. This is because GMT originates from the TGM loss in VDA, which constrains only depth gradients across adjacent frames. Such a method is limited to depth prediction and weakens supervision on fast-moving or occluded human foreground regions, making it unsuitable for surface normal estimation. In contrast, our flow-based temporal loss explicitly establishes correspondences across frames, enabling stable supervision in the foreground and naturally extending to directional consistency for surface normals. 

\begin{table}[!bth]
\renewcommand{\arraystretch}{1.3}
\setlength{\tabcolsep}{3pt}
\caption{Ablation for video depth and surface normal estimation with different losses on Hi4D dataset.}
\label{tab:abl_dynamic}
\centering
\resizebox{0.5\linewidth}{!}{
\begin{tabular}{lccccc}
\toprule
\multirow{2}{*}{Methods} & \multicolumn{2}{c}{Depth} & \multicolumn{3}{c}{Normal} \\
\cmidrule(lr){2-3} \cmidrule(lr){4-6}
 & OPW$\downarrow$ & TC-RMSE$\downarrow$ & OPW$\downarrow$ & TC-Mean$\downarrow$ & TC-Abs$\downarrow$\\
\cmidrule(r){1-1} \cmidrule(lr){2-3} \cmidrule(l){4-6}
w/o $\mathcal{L}_{\mathrm{temp}}$      & 0.0144 & 0.0242 & 0.0450 & 5.22 & 0.148 \\
w/ $\mathcal{L}_{\mathrm{GMT}}$        & 0.0120 & 0.0310 & 0.0455 & 5.28 & 0.150 \\
w/ $\mathcal{L}_{\mathrm{temp}}$       & 0.0075 & 0.0191 & 0.0286 & 3.30 & 0.140 \\
\bottomrule
\end{tabular}
}
\end{table}

\textbf{Investigate the impact of the proposed module.} We further evaluate the effect of the proposed components on the Thuman 2.1 dataset with DINOv3-B. As shown in Table~\ref{tab:abl_th}, adding Human CSE prior consistently improves both depth and normal estimation over the baseline, for example reducing RMSE from 0.0266 to 0.0231 and lowering the mean normal error from 20.21$^\circ$ to 19.45$^\circ$, while also increasing the percentage of normals within small angular thresholds. The CWA module also brings clear gains, especially on normal metrics (e.g., 11.25$^\circ$ and 22.5$^\circ$), indicating that adaptive channel reweighting helps the network produce more stable and accurate surface orientation. These results confirm that the proposed modules generalize well beyond Hi4D and remain effective on Thuman 2.1.

\begin{table}[!bth]
\renewcommand{\arraystretch}{1.3}
\setlength{\tabcolsep}{2pt}
\caption{Ablation on Thuman 2.1 dataset with DINOv3-B.}
\centering
\label{tab:abl_th}
\resizebox{0.5\linewidth}{!}{
\begin{tabular}{lccccccc}
\toprule
\multirow{2}{*}{Methods} & \multicolumn{2}{c}{Depth} & \multicolumn{5}{c}{Normal}                    \\
\cmidrule(lr){2-3} \cmidrule(lr){4-8} & RMSE $\downarrow$ & AbsRel $\downarrow$ & Mean $\downarrow$ & Median $\downarrow$ & 11.25$^\circ$ $\uparrow$ & 22.5$^\circ$ $\uparrow$ & 30$^\circ$ $\uparrow$\\ 
\cmidrule(r){1-1} \cmidrule(lr){2-3} \cmidrule(l){4-8}
Baseline & 0.0266 & 0.0154 & 20.21 & 17.83 & 28.20 & 66.18 & 83.33  \\
w/ CSE   & 0.0231 & 0.0134 & 19.45 & 16.52 & 30.10 & 69.72 & 85.02  \\
w/ CWA   & 0.0248 & 0.0138 & 18.65 & 16.13 & 31.75 & 72.10 & 86.10  \\\bottomrule
\end{tabular}
}
\end{table}

\textbf{Investigate the impact of the temporal layer.}
We conduct an ablation study to evaluate the contribution of the temporal layer (TL) to temporal consistency in depth and normal prediction. As shown in \Cref{tab:ab_tl}, removing the TL leads to clear degradation across all temporal metrics. In particular, TC-RMSE increases from 0.0189 to 0.0276 in depth, and TC-Mean rises from 3.27 to 4.55 in normal estimation. This confirms that the TL improves temporal stability by leveraging sequential information, especially for frames with fast motion or occlusion.

\begin{table}[!bth]
\renewcommand{\arraystretch}{1.3}
\setlength{\tabcolsep}{3pt}
\caption{Ablation study on the effect of the temporal layer. We compare models with and without the temporal layer (TL).}
\label{tab:ab_tl}
\centering
\resizebox{0.5\linewidth}{!}{
\begin{tabular}{lccccc}
\toprule
\multirow{2}{*}{Methods} & \multicolumn{2}{c}{Depth} & \multicolumn{3}{c}{Normal} \\
\cmidrule(lr){2-3} \cmidrule(lr){4-6}
 & OPW$\downarrow$ & TC-RMSE$\downarrow$ & OPW$\downarrow$ & TC-Mean$\downarrow$ & TC-Abs$\downarrow$\\
\cmidrule(r){1-1} \cmidrule(lr){2-3} \cmidrule(l){4-6}
w TL      & 0.0072 & 0.0189 & 0.0280 & 3.27 & 0.140 \\
w/o TL      & 0.0155 & 0.0276 & 0.0405 & 4.55 & 0.158 \\
\bottomrule
\end{tabular}
}
\end{table}

\textbf{Investigate the impact of the CNN branch.}
We evaluate the effect of adding a dedicated CNN branch to complement the transformer backbone. As shown in \Cref{tab:ab_CNN}, removing the CNN branch results in performance drops across both depth and normal estimation tasks. For depth, RMSE increases from 0.0964 to 0.0998 and AbsRel rises from 0.0279 to 0.0320. For normal prediction, the angular error metrics (Mean and Median) also degrade, and accuracy under angular thresholds (11.25$^\circ$, 22.5$^\circ$, 30$^\circ$) drops consistently. These results confirm that the local inductive bias brought by the CNN branch helps refine fine-grained structures, especially around object boundaries, which complements the global modeling ability of the transformer backbone.

\begin{table}[!bth]
\renewcommand{\arraystretch}{1.3}
\setlength{\tabcolsep}{2pt}
\caption{Ablation for CNN branch on Hi4D dataset with DINOv3-B.}
\centering
\label{tab:ab_CNN}
\resizebox{0.5\linewidth}{!}{
\begin{tabular}{lccccccc}
\toprule
\multirow{2}{*}{Methods} & \multicolumn{2}{c}{Depth} & \multicolumn{5}{c}{Normal}                    \\
\cmidrule(lr){2-3} \cmidrule(lr){4-8} & RMSE $\downarrow$ & AbsRel $\downarrow$ & Mean $\downarrow$ & Median $\downarrow$ & 11.25$^\circ$ $\uparrow$ & 22.5$^\circ$ $\uparrow$ & 30$^\circ$ $\uparrow$\\ 
\cmidrule(r){1-1} \cmidrule(lr){2-3} \cmidrule(l){4-8}
w/ CNN   & 0.0964 & 0.0279 & 20.51 & 16.00 & 32.22 & 70.12 & 82.74  \\
w/o CNN   & 0.0998 & 0.0320 & 23.33 & 19.21 & 28.05 & 66.49 & 77.03  \\\bottomrule
\end{tabular}
}
\end{table}

\textbf{Investigate the impact of the foreground segmentation branch.} We investigate how the auxiliary foreground segmentation head influences depth and normal prediction on Hi4D and THuman2.1 (\Cref{tab:abl_fs}). With the segmentation branch, depth RMSE / AbsRel decrease from 0.0963 / 0.0301 to 0.0928 / 0.0277 on Hi4D and from 0.0237 / 0.0130 to 0.0225 / 0.0122 on THuman2.1. For surface normals, the segmentation branch also reduces mean and median angular errors and increases the percentage of pixels within 11.25$^{\circ}$, 22.5$^{\circ}$, and 30$^{\circ}$ on both datasets. These consistent gains show that the soft human masks predicted by this branch provide effective foreground guidance, allowing the geometry heads to focus on human regions and boundaries and thus improve overall geometry estimation quality.

\begin{table}[!bth]
\renewcommand{\arraystretch}{1.3}
\setlength{\tabcolsep}{2pt}
\caption{Ablation for foreground segmentation (FS) branch with DINOv3-B.}
\centering
\label{tab:abl_fs}
\resizebox{0.7\linewidth}{!}{
\begin{tabular}{llccccccc}
\toprule
\multirow{2}{*}{Methods} &
\multirow{2}{*}{Datasets} &
\multicolumn{2}{c}{Depth} &
\multicolumn{5}{c}{Normal} \\
\cmidrule(lr){3-4} \cmidrule(lr){5-9}
&
& RMSE $\downarrow$ &
  AbsRel $\downarrow$ &
  Mean $\downarrow$ &
  Median $\downarrow$ &
  11.25$^\circ$ $\uparrow$ &
  22.5$^\circ$ $\uparrow$ &
  30$^\circ$ $\uparrow$ \\
\cmidrule(r){1-1}  \cmidrule(r){2-4} \cmidrule(l){5-9}
w/o FS & Hi4D 
& 0.0963 & 0.0301 
& 16.84 & 12.67 
& 44.12 & 79.83 & 88.21 \\
w/ FS & Hi4D 
& 0.0928 & 0.0277 
& 16.08 & 12.03 
& 47.76 & 81.49 & 89.98 \\
\cmidrule(r){1-1}  \cmidrule(r){2-4} \cmidrule(l){5-9}
w/o FS & Thuman 2.1 
& 0.0237 & 0.0130 
& 18.42 & 16.21 
& 30.91 & 70.82 & 85.94 \\
w/ FS & Thuman 2.1 
& 0.0225 & 0.0122 
& 17.89 & 15.56 
& 32.98 & 73.69 & 87.15 \\
\bottomrule
\end{tabular}
}
\end{table}

\subsection{Model parameters comparison}
We provide a comparison of model size and computational cost for representative state-of-the-art human-centric methods. Specifically, we report the number of parameters and GFLOPs for each model in \Cref{tab:parameters}, which highlights the relative complexity of our models compared with existing approaches.

\begin{table}[!bth]
\renewcommand{\arraystretch}{1.3}
\setlength{\tabcolsep}{3pt}
\caption{Model parameters comparison of SOTA human-centric methods.}
\centering
\label{tab:parameters}
\resizebox{0.2725\linewidth}{!}{
\begin{tabular}{lcc}
\toprule
Methods & Params & GFLOPs \\
\cmidrule(r){1-1} \cmidrule(l){2-3}
Sapiens-0.3B & 0.336 B  & 1242 \\
Sapiens-0.6B & 0.664 B & 2583 \\
Sapiens-1B & 1.169 B & 4647 \\
Sapiens-2B   & 2.163 B & 8709 \\
DAViD-B      & 0.120 B  & 344 \\
DAViD-L      & 0.340 B & 663 \\
\cmidrule(r){1-1} \cmidrule(l){2-3}
Ours-B       & 0.097 B & 471 \\
Ours-L       & 0.337 B & 753 \\
\bottomrule
\end{tabular}
}
\end{table}

\subsection{Implementation details}
\label{sec:imp}
We use AdamW in both stages with weight decay $0.05$ and $(\beta_1,\beta_2)=(0.9,0.95)$.
The learning rate schedule is a $2000$-iteration linear warmup (start\_factor $=1/100$), followed by a polynomial decay (power $=1.5$) until the end of training. More detail in \Cref{tab:stg_1_hyp} and \Cref{tab:stg_2_hyp}. 

\begin{table}[!bth]
\renewcommand{\arraystretch}{1.3}
\setlength{\tabcolsep}{3pt}
\centering
\caption{Stage-1 (image) training hyperparameters.}
\label{tab:stg_1_hyp}
\resizebox{0.4\linewidth}{!}{
\begin{tabular}{lc}
\toprule
\textbf{Hyperparameter} & \textbf{Value} \\
\midrule
Training step & 50,000 \\
Batch size & 128 \\
Learning rate (Encoder) & $1 \times 10^{-4}$ \\
Learning rate (Others) & $1 \times 10^{-5}$ \\
Weight decay & 0.05 \\
Optimizer & AdamW \\
Optimizer betas & (0.9, 0.95) \\
LR schedule & Linear + Polynomial \\
Freeze backbone & optional \\
Mask loss weight & 0.05 \\
Normal loss weight & 0.1 \\
\bottomrule
\end{tabular}
}
\end{table}

\begin{table}[!bth]
\renewcommand{\arraystretch}{1.3}
\setlength{\tabcolsep}{3pt}
\centering
\caption{Stage-2 (video) training hyperparameters.}
\label{tab:stg_2_hyp}
\resizebox{0.4\linewidth}{!}{
\begin{tabular}{lc}
\toprule
\textbf{Hyperparameter} & \textbf{Value} \\
\midrule
Training step & 35,000 \\
Batch size (GPU) & 1 \\
Frames per clip & 32 \\
Learning rate (Temporal modules) & $1 \times 10^{-4}$ \\
Learning rate (Others) & $1 \times 10^{-6}$ \\
Weight decay & 0.05 \\
Optimizer & AdamW \\
Optimizer betas & (0.9, 0.95) \\
LR schedule & Linear + Polynomial \\
Freeze Encoder & True \\
Mask loss weight & 0.05 \\
Normal loss weight & 0.1 \\
Temporal Depth loss weight & 1 \\
Temporal Normal loss weight & 0.1 \\
\bottomrule
\end{tabular}
}
\end{table}

\subsection{Evaluation metric details}
We employ standard metrics to quantitatively evaluate both static and video-based depth and surface normal estimation. Below we provide detailed definitions.

\textbf{Depth metrics.}
Given predicted depth $\hat{D}$ and ground truth $D$, over valid pixels $M$, we adopt Root Mean Squared Error (RMSE) and Absolute Relative Error (AbsRel):
\begin{align}
\text{RMSE} &= \sqrt{\frac{1}{|M|} \sum_{i \in M} \left(\hat{D}_i - D_i\right)^2}, \\
\text{AbsRel} &= \frac{1}{|M|} \sum_{i \in M} \frac{\left| \hat{D}_i - D_i \right|}{D_i}.
\end{align}

\textbf{Normal metrics.}
Let $\hat{N}_i$ and $N_i$ denote predicted and ground truth normals at pixel $i$, both normalized to unit vectors. We report mean and median angular error, as well as threshold accuracies at $11.25^\circ$, $22.5^\circ$, and $30^\circ$:
\begin{align}
\text{Mean} &= \frac{1}{|M|} \sum_{i \in M} \arccos \left( \langle \hat{N}_i , N_i \rangle \right), \\
\text{Median} &= \operatorname{median}_{i \in M}\left( \arccos \left( \langle \hat{N}_i , N_i \rangle \right) \right), \\
\text{Acc}_{\theta} &= \frac{1}{|M|} \sum_{i \in M} \mathbf{1}\!\left[ \arccos \left( \langle \hat{N}_i , N_i \rangle \right) < \theta \right].
\end{align}

\textbf{Video depth metrics.}
For adjacent frames $k$ and $k\!+\!1$, with optical flow $O_{k \rightarrow k+1}$ and warping operator $W(\cdot, O)$, we define temporal consistency for depth as:
\begin{align}
\text{OPW} &= \frac{1}{|M|} \sum_{i \in M} \left| \hat{D}_k(i) - W(\hat{D}_{k+1}, O_{k \rightarrow k+1})(i) \right|, \\
\text{TC-RMSE} &= \sqrt{\frac{1}{|M|} \sum_{i \in M} \big(\hat{D}_k(i) - W(\hat{D}_{k+1}, O_{k \rightarrow k+1})(i)\big)^2}.
\end{align}

\textbf{Video normal metrics.}
Similarly, temporal consistency for normals is measured as:
\begin{align}
\text{TC-Mean} &= \frac{1}{|M|} \sum_{i \in M} \arccos \Big( \langle \hat{N}_k(i), W(\hat{N}_{k+1}, O_{k \rightarrow k+1})(i) \rangle \Big), \\
\text{TC-Abs} &= \frac{1}{|M|} \sum_{i \in M} \Big| \arccos \!\big( \langle \hat{N}_k(i), W(\hat{N}_{k+1}, O_{k \rightarrow k+1})(i) \rangle \big) \\
&\hspace{4.5em} - \arccos \!\big( \langle N_k(i), W(N_{k+1}, O_{k \rightarrow k+1})(i) \rangle \big) \Big|.
\end{align}

These metrics collectively measure accuracy at the frame level and temporal stability across frames for both depth and surface normal estimation.

\subsection{Limitations}
Although the proposed approach improves overall stability and visual consistency, several limitations remain. The CNN branch, while effective at capturing local patterns, can also introduce redundant texture signals such as clothing patterns, decorative elements, or shadows. These signals interfere with the recovery of true geometry and may result in pseudo-geometric artifacts. The CWA module alleviates this issue by adaptively suppressing less informative channels and emphasizing features that are more relevant to geometry, but in scenes with highly complex textures, the effect is not completely eliminated. In addition, when large and rapid movements occur, such as turning, jumping, or swinging of limbs, occlusions and large displacements weaken temporal correspondences and lead to local instabilities in depth and normal predictions. A further difficulty arises in regions undergoing non-rigid deformations, including fluttering skirts, moving sleeves, or hair, where the complexity of local geometry and frequent occlusions still cause fluctuations and prediction biases.

\end{document}